\documentclass{article}

\usepackage{microtype}
\usepackage{graphicx}
\usepackage{subcaption}
\usepackage{booktabs}
\usepackage{xspace}
\usepackage{hyperref}
\usepackage{multirow}
\usepackage[table]{xcolor}
\usepackage{tabularx}
\usepackage{array}
\usepackage{listings}
\usepackage{tcolorbox}
\usepackage{enumitem}
\usepackage{pifont}
\usepackage{placeins}

\usepackage[preprint]{icml2026}

\usepackage{amsmath}
\usepackage{amssymb}
\usepackage{mathtools}
\usepackage{amsthm}

\usepackage[capitalize,noabbrev]{cleveref}

\theoremstyle{plain}

\theoremstyle{definition}

\theoremstyle{remark}

\usepackage[disable,textsize=tiny]{todonotes}

\icmltitlerunning{Find, Fix, Reason: Context Repair for Video Reasoning}

\makeatletter
\newif\if@firstpa
\renewcommand{\icmlauthor}[2]{%
  \ificmlshowauthors
    \mbox{\bf #1}\,\@firstpatrue\@for\theaffil:=#2\do{\@pa{\theaffil}}%
    \addtofullauthorlist{#1}%
  \else
    \ifdefined\@icmlfirsttime\else
      \gdef\@icmlfirsttime{1}
      \mbox{\bf Anonymous Authors}\@pa{@anon} \addtofullauthorlist{Anonymous Authors}
    \fi
  \fi
}
\renewcommand{\@pa}[1]{%
  \ifcsname the@affil#1\endcsname
  \else
    \ifcsname @icmlsymbol#1\endcsname
    \else
      \stepcounter{@affiliationcounter}%
      \newcounter{@affil#1}%
      \setcounter{@affil#1}{\value{@affiliationcounter}}%
    \fi
  \fi%
  \if@firstpa\@firstpafalse\else\textsuperscript{,}\fi
  \ifcsname @icmlsymbol#1\endcsname
    \textsuperscript{\csname @icmlsymbol#1\endcsname}%
  \else
    \textsuperscript{\arabic{@affil#1}}%
  \fi
}
\makeatother

\newcommand{\ourrepo}{{\fontfamily{lmtt}\selectfont FFR}\xspace}
\newcommand{\mllms}{{MLLMs}\xspace}
\newcommand{\mlrms}{{MLRMs}\xspace}

\newcommand{\chainofthought}{{{CoT}}\xspace}

\newcommand{\eg}{\textit{e.g.,}\xspace}
\newcommand{\etc}{\textit{etc.}\xspace}
\newcommand{\ie}{\textit{i.e.,}\xspace}

\begin{document}

\twocolumn[
  \icmltitle{Find, Fix, Reason: Context Repair for Video Reasoning}

  \icmlsetsymbol{equal}{*}

  \begin{icmlauthorlist}
    \icmlauthor{Haojian Huang}{hkust,knowin,equal}
    \icmlauthor{Chuanyu Qin}{iie,equal}
    \icmlauthor{Yinchuan Li}{knowin}
    \icmlauthor{Yingcong Chen}{hkust,knowin}
  \end{icmlauthorlist}

  \icmlaffiliation{hkust}{The Hong Kong University of Science and Technology (Guangzhou)}
  \icmlaffiliation{iie}{Institute of Information Engineering, Chinese Academy of Sciences}
  \icmlaffiliation{knowin}{Knowin AI}

  \icmlcorrespondingauthor{Haojian Huang}{haojianhuang927@gmail.com}
  \icmlcorrespondingauthor{Yingcong Chen}{yingcongchen@hkust-gz.edu.cn}

  \icmlkeywords{Video Reasoning, Reinforcement Learning, Multi-modal Large Language Models}

  \vskip 0.3in
]

\makeatletter\let\Hy@OrigWarning\Hy@Warning\renewcommand{\Hy@Warning}[1]{}\makeatother
\printAffiliationsAndNotice{\icmlEqualContribution}
\makeatletter\let\Hy@Warning\Hy@OrigWarning\makeatother
\begin{abstract}
Reinforcement learning has advanced video reasoning in large multi-modal models, yet dominant pipelines either rely on on-policy self-exploration, which plateaus at the model’s knowledge boundary, or hybrid replay that mixes policies and demands careful regularization. Dynamic context methods zoom into focused evidence but often require curated pretraining and two-stage tuning, and their context remains bounded by a small model’s capability. In contrast, larger models excel at instruction following and multi-modal understanding, can supply richer context to smaller models, and rapidly zoom in on target regions via simple tools. Building on this capability, we introduce an observation-level intervention: a frozen, tool-integrated teacher identifies the missing spatiotemporal dependency and provides a minimal evidence patch (\eg timestamps, regions \etc) from the original video while the question remains unchanged. The student answers again with the added context, and training updates with a chosen-rollout scheme integrated into Group Relative Policy Optimization (GRPO). We further propose a Robust Improvement Reward (RIR) that aligns optimization with two goals: outcome validity through correct answers and dependency alignment through rationales that reflect the cited evidence. Advantages are group-normalized across the batch, preserving on-policy exploration while directing it along causally meaningful directions with minimal changes to the training stack. Experiments on various related benchmarks show consistent accuracy gains and strong generalization.
Code is available at \href{https://github.com/JethroJames/FFR.git}{\textcolor{magenta}{\ourrepo}}.
\end{abstract}    
\begin{figure}[!t]
    \centering
    \includegraphics[width=0.96\linewidth]{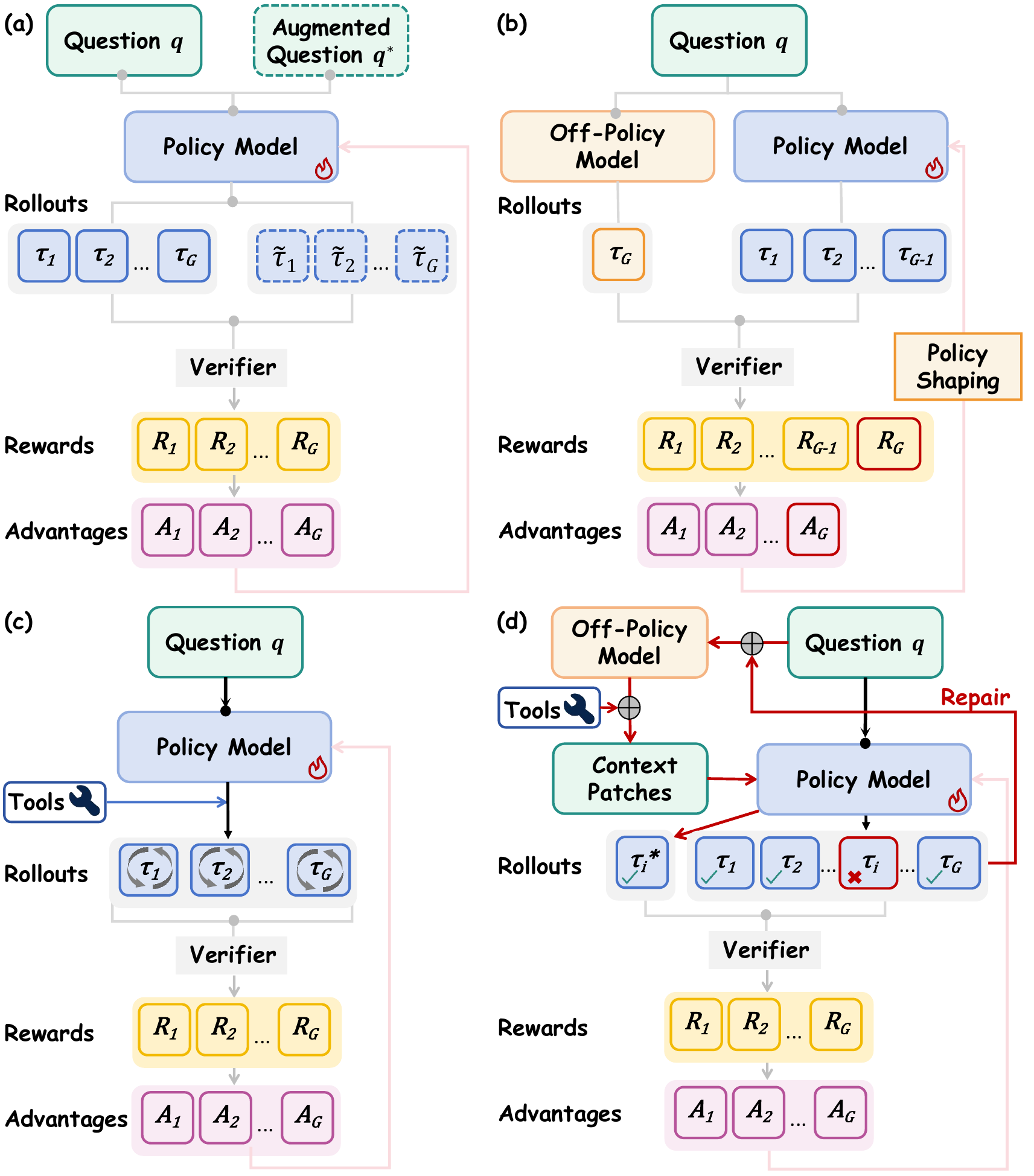}
\caption{\textbf{Comparison of video reasoning training regimes.} 
\textbf{\textit{(a)} On-policy:} relies on self-exploration and data scaling \citep{VideoR1,VideoRFT,Time-R1}. 
\textbf{\textit{(b)} Hybrid:} blends buffered trajectories via policy shaping. 
\textbf{\textit{(c)} Tool-use:} performs multi-round, budgeted context retrieval. 
\textbf{\textit{(d)} Ours:} a frozen teacher repairs failed rollouts with minimal patches; the policy then re-answers and updates on the rectified trajectories.}
    \label{fig:comp}
    \vspace{-1.5em}
\end{figure}
\section{Introduction}
\label{sec:intro}

Recent advancements in rule-based Reinforcement Learning (RL) have significantly enhanced the reasoning capabilities of Multi-modal Large Language Models (MLLMs), particularly in video understanding~\citep{VideoR1,VideoChatR1,TinyLLaVA-Video-R1,zhang2025thinking}. Unlike mathematical~\citep{zhan2025exgrpo,yan2025learning} or code-based tasks~\citep{ding2024semcoder,du2025dependeval,ni2024next,zhang2024codeagent}, video reasoning must interpret dynamic spatiotemporal cues to generate causally grounded explanations~\citep{huang2025demr,chen2024finecliper,hu2024multimodal,li2025scagiqa}. However, progress is hindered by the lack of reasoning-rich corpora, preventing systematic exploration of model robustness across diverse scenarios.

Existing GRPO-based frameworks~\citep{VideoR1,VideoChatR1,Time-R1,zhang2025rewatch} rely chiefly on on-policy self-exploration (\textbf{Figure~\textcolor{magenta}{\ref{fig:comp}} (\textit{a})}), which often hits an exploration ceiling by reinforcing existing inductive biases. While injecting off-policy guidance from stronger teachers (\textbf{Figure~\textcolor{magenta}{\ref{fig:comp}} (\textit{b})}) can stabilize convergence~\citep{On-policyDistillation,yan2025learning}, it frequently leads to entropy collapse. Furthermore, recent tool-integrated reasoners~\citep{hong2025deepeyesv2,zhang2025thyme,su2025pixel,wang2025video,he2025framethinker,lai2025mini,su2025thinking,li2025select,zheng2024videogenofthought} attempt to focus on critical evidence through multi-round context search (\textbf{Figure~\textcolor{magenta}{\ref{fig:comp}} (\textit{c})}), but remain bounded by the smaller model's own capability, leading to repetitive "self-doubt" cycles and requiring costly two-stage training~\citep{su2025pixel,he2025framethinker}.

To bypass these constraints, we propose \textbf{Find, Fix, Reason (FFR)} (\textbf{Figure~\textcolor{magenta}{\ref{fig:comp}} (\textit{d})}), which leverages a frozen, tool-integrated teacher for observation-level intervention. Instead of relying on the student's self-search, the teacher diagnoses missing spatiotemporal dependencies in failed rollouts and supplies a \textit{minimal evidence patch} (\eg timestamps, regions) while the question remains unchanged. This patch acts as a controlled causal intervention, suppressing spurious shortcuts and promoting dependency-grounded reasoning. 

We integrate this mechanism into GRPO via a \textbf{Robust Improvement Reward (RIR)} that targets outcome validity and dependency alignment. Since patches are triggered only on failures, the curriculum adaptively improves temporal coherence and explanation fidelity. Our contributions are:
\vspace{-5pt}
\begin{itemize}
  \setlength{\itemsep}{0pt}
  \setlength{\parsep}{0pt}
  \item We use a frozen teacher to provide minimal patches that rectify missing dependencies while remaining on-policy nature.
  \item We design a RIR that jointly optimizes outcome validity and the reasoning process.
  \item Evaluations confirm FFR's effectiveness, showing that smaller student models can match or even surpass larger teachers.
\end{itemize}
\vspace{-5pt}
\section{Related Work}
\label{sec:related_work}
\subsection{Multimodal Large Reasoning Models}
Multimodal reasoning with \mllms, namely multimodal large reasoning models (\mlrms) have evolved from \emph{modular prompting} to \emph{rule-based RL}. Early approaches decouple perception from inference—visual facts are distilled into textual hints, then consumed by a separate reasoning stage~\citep{MMCoT,DDCoT,li2024enhancing,huang2024crest,huang2025tuned}; video counterparts further impose stage-wise templates and external memory/tool use to scaffold multi-step inference~\citep{AoTD,VoT,STEP,DoraemonGPT,chen2025finequest}. This structured supervision stabilizes training but tends to overfit handcrafted stages, yielding brittle generalization to temporal and causal patterns. More recently, rule-based RL~\citep{guo2025deepseek,o1,kimik1.5,qu2025survey} has shifted the focus from hand-crafted pipelines to \emph{programmatic, verifiable rewards}, encouraging \chainofthought (chain-of-thought)-style behavior in code~\citep{robeyns2025improving,pennino2025reasoning}, math~\citep{yan2025learning,zhan2025exgrpo}, and images~\citep{VisualRFT,R1VL,ReasonRFT,VisionR1,R1-OneVision}, with extensions to videos~\citep{VideoR1,VideoRFT,VideoChatR1,TinyLLaVA-Video-R1}. In this Reinforcement Learning with Verifiable Rewards (RLVR) paradigm, the model outputs a canonicalized answer; a rule/regex validator exact-matches it to a gold label to yield a binary reward that deters reward hacking and directly drives subsequent advantage estimation for scalable training. For video tasks, recent variants~\citep{chen2025versavid,Time-R1,Open-o3,zhang2025rewatch} broaden subtasks coverage and data, refine credit assignment (\eg, difficulty-aware~\citep{DeepVideo-R1} and token-level importance-based~\citep{dang2025reinforcing}), and scale test-time compute~\citep{Video-rts} to better steer single-model trajectories. Yet more signals or compute hardly internalize \emph{correct reasoning dependencies}~\citep{rl_limit,zhao2025echo}. Thus, we introduce a teacher model that debiases spurious dependencies in the student's reasoning, turning guidance into robust, causally grounded gains.

\subsection{On-Policy and Off-Policy RL for Video QA}
Reinforcement learning (RL) for Video QA has evolved into two primary paradigms based on how experiences are utilized during policy updates. \emph{On-policy} methods rely on trajectories generated by the current policy, updating immediately. This approach ensures stable improvements, especially when rewards are verifiable and closely tied to the answer formats, enabling effective exploration. However, it can be data and compute intensive. Several large video models~\citep{VideoR1, VideoRFT, ReWatch-R1, Time-R1} follow this route and have demonstrated consistent performance gains.
In contrast, \emph{off-policy} methods, such as DPO~\citep{rafailov2023direct} and its variants, leverage historical trajectories or preferences generated by older or broader policies, improving sample efficiency and accelerating performance gains~\citep{zhang2024direct, huang2025vistadpo, chen2025physhpo, ahn2025isr, dahal2025povqa, zhao2026star, zhao2025evoempirbench}. However, off-policy learning introduces the challenge of distribution shift~\citep{zhan2025exgrpo, yan2025learning}.

Building on progress in mathematical reasoning, several video off-policy RL frameworks adopt hybrid policies that mix replay buffers with on-policy samples, using past stable traces to aid current learning~\citep{On-policyDistillation,yan2025learning,zhang2025critique,kulkarni2025avatar}. Yet replay-based schemes depend on careful replay management and mixed-policy regularizer~\citep{chen2024bovila}. Without them, exploration shrinks and reward hacking emerges. A parallel line zooms into dynamically acquired, dependency-focused context to boost reasoning, but typically requires curated pretraining and two-stage fine-tuning~\citep{he2025framethinker,zhang2025thinking,wang2025video}, yet they still rely on context retrieved within a small model's capability envelope, limiting the potential to overcome the performance bottleneck. To overcome these limits, we introduce a larger, tool-integrated teacher that detects and challenges spurious or missing dependencies, delivering targeted evidence and prompts that steer exploration and refine the student's internal reasoning.

\section{Methodology}
\label{sec:methodology}
\subsection{Preliminary}

\textbf{Group Relative Policy Optimization (GRPO).}
GRPO~\cite{guo2025deepseek} can be viewed as a \emph{value-free, group-contrastive} policy gradient method tailored to verifiable-reward RL. Rather than training a separate value function, GRPO converts a \emph{set} of $G$ candidate solutions sampled for the \emph{same} query into a normalized advantage signal. This per-query normalization makes the update insensitive to reward scale or shift across different queries and fits naturally with binary RLVR rewards~\cite{guo2025deepseek,zeng2025simplerl,drgrpo}.

\textbf{Setup.}
Let $\pi_{\theta_{\text{old}}}$ denote the policy that generates samples and $\pi_{\theta}$ the policy being optimized. For a question $q$, draw $G$ solutions $\{\tau_i\}_{i=1}^G \sim \pi_{\theta_{\text{old}}}$ and compute rewards $R(\tau_i)$ with a verifiable checker. GRPO forms a \emph{group-relative} advantage by standardizing rewards within the set:
\begin{equation}
A_i \;=\; \frac{R(\tau_i) - \mathrm{mean}\big(\{R(\tau_j)\}_{j=1}^G\big)}{\mathrm{std}\big(\{R(\tau_j)\}_{j=1}^G\big)+\delta}.
\label{eq:grpo-adv}
\end{equation}
This broadcasts a single sequence-level $A_i$ to all token steps of $\tau_i$ and serves as a variance-reduced baseline without a learned critic. $\delta$ is a small constant
for numerical stability.

\textbf{Objective.}
GRPO adopts the clipped surrogate from PPO~\cite{PPO} with token-wise importance ratios
\(
r_{i,t}(\theta)=\frac{\pi_{\theta}(\tau_{i,t}\mid q,\tau_{i,<t})}{\pi_{\theta_{\text{old}}}(\tau_{i,t}\mid q,\tau_{i,<t})}
\)
derived from policy gradient theory~\cite{policy_gradient}:
\begin{equation}
\begin{split}
\mathcal{J}_{\mathrm{GRPO}}(\theta)
&= \frac{1}{\sum_{i=1}^G |\tau_i|}\,
\sum_{i=1}^G \sum_{t=1}^{|\tau_i|}
\mathrm{CLIP}\!\big(r_{i,t}(\theta),A_i,\epsilon\big) \\
&\quad - \beta\,\mathbb{D}_{\mathrm{KL}}\!\big[\pi_{\theta}\,\|\,\pi_{\mathrm{ref}}\big].
\end{split}
\label{eq:grpo-obj}
\end{equation}

where $\mathrm{CLIP}(r,A,\epsilon)\triangleq \min\!\big(r\cdot A,\;\mathrm{clip}(r;1-\epsilon,1+\epsilon)\cdot A\big)$ and $\mathbb{D}_{\mathrm{KL}}$ regularizes updates toward a reference model~\cite{trpo}. The KL term can be tuned or omitted in practice~\cite{prime,yu2025dapo,hu2025open}, but the method remains effectively on-policy since samples are drawn from $\pi_{\theta_{\text{old}}}$, a close predecessor of $\pi_{\theta}$.

\textbf{Discussion.}
By replacing a learned critic with group normalization in Eq.~\eqref{eq:grpo-adv}, GRPO scales cleanly under RLVR, where rewards are often sparse or binary, and has shown strong results across tasks~\cite{guo2025deepseek,zeng2025simplerl,drgrpo}. The contrastive, per-query normalization sharpens credit assignment among competing solutions to the same $q$, while the PPO-style clipping in Eq.~\eqref{eq:grpo-obj} stabilizes token-level likelihood updates.

\textbf{RLVR in Video Understanding and Reasoning.}
Given a video-question-answer triplet $(v, q, y) \sim \mathcal{D}$ sampled from the data distribution $\mathcal{D}$, where $v$ is the video, $q$ is the question, and $y$ is the ground-truth answer, the task is to generate a reasoning trajectory $\tau = (\tau_1, \ldots, \tau_G) \sim \pi_\theta(\cdot | v, q)$ from a policy model parameterized by $\theta$. The final answer $\hat{y}$ is extracted from the reasoning trajectory $\tau_i$ using a parser $\phi$, \ie, $\hat{y_i} = \phi(\tau_i)$. A verifier $\mathcal{V}$ checks the correctness of the answer by comparing $\hat{y}$ with the ground-truth $y$, yielding a binary reward:
\begin{equation}
R(\tau_i) = \mathcal{V}(\hat{y_i}, y) \in \{0, 1\}.
\end{equation}
This reward design minimizes reward hacking~\cite{reward_hack}, enabling successful RL training scaling~\cite{guo2025deepseek}.

\begin{figure*}[!t]
    \centering
    \includegraphics[width=0.96\linewidth]{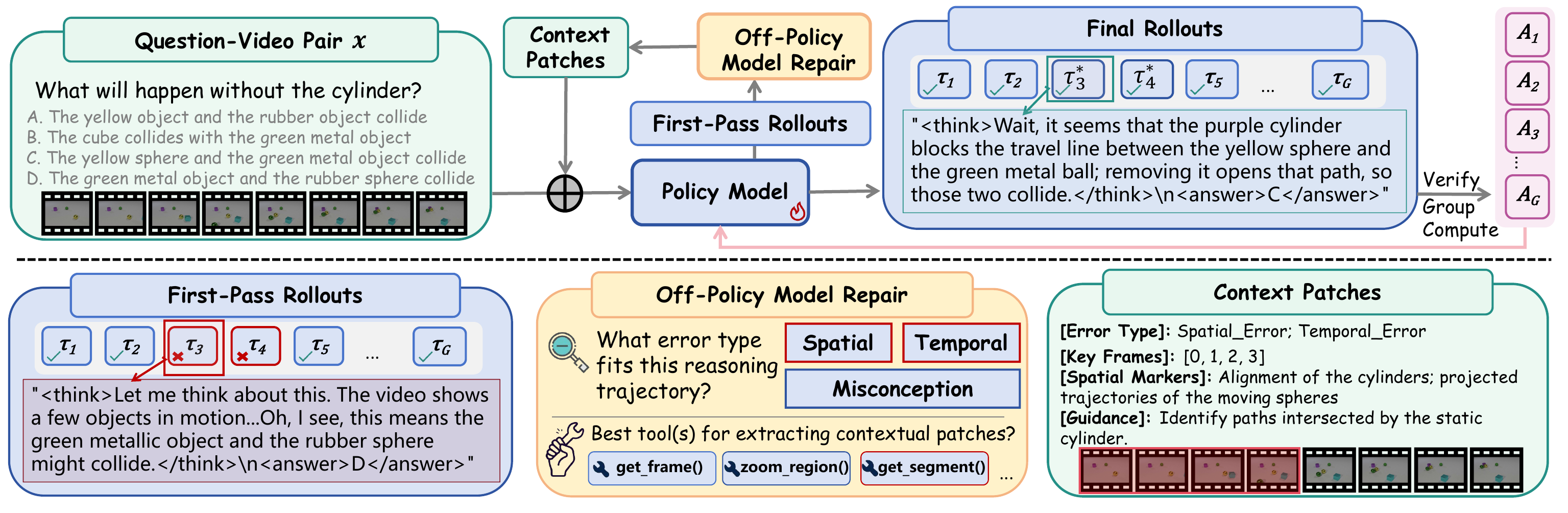}
\caption{Overview of FFR. Given a video–question pair, the policy model generates first-pass rollouts and a verifier scores them. For failures, a frozen, tool-integrated teacher diagnoses the missing spatiotemporal dependency and extracts a minimal evidence patch from the same video (\eg frames or segments); the policy then re-answers the same question with this patch to produce a repaired rollout. Group-normalized rewards drive a GRPO update on the chosen rollout only (original if correct, otherwise repaired), which preserves on-policy learning, avoids costly multi-round search, and steers exploration toward evidence-aligned reasoning.}
    \label{fig:flowchart}
\vspace{-1em}
\end{figure*}

\subsection{Find, Fix, Reason: Context Repair for Video Reasoning}
\textbf{Observations and Motivations.}
On-policy RL encourages self-exploration but often saturates at the model’s knowledge boundary, limiting gains~\citep{rl_limit,zhao2025echo}. In math reasoning, off-policy trajectories can guide learning~\citep{yan2025learning,prime,mroueh2025revisiting,On-policyDistillation}, yet they introduce hand-crafted shaping and reliance on teacher policies. The challenge is sharper in video reasoning: multi-modal context creates many candidate dependencies, so fitting off-policy traces risks overfitting to incidental textual, temporal, or spatial cues. Unlike math—with relatively stable thought templates that enable stepwise critique~\citep{yang2025reasonflux,yang2024buffer,zhang2025critique,yang2024supercorrect}—video lacks a canonical procedure, raising supervision cost and inviting reward hacking; simple self-replay recycles these issues~\citep{zhan2025exgrpo,li2025repo,kulkarni2025avatar}. Recent tool-integrated methods perform multi-round context retrieval to surface dependencies~\citep{su2025pixel,wang2025video,su2025thinking,zhang2025thinking,he2025framethinker}, but typically rely on small models with curated pretraining and two-stage fine-tuning, which hinders rapid deployment and limits their ability to break the capability ceiling. In contrast, larger models have stronger instruction following and tool use, and can more reliably localize accurate, task relevant context. This motivates a question: \emph{can off-policy signals from a larger, tool-integrated teacher reveal the essential dependencies and target weaknesses in on-policy rollouts, thereby focusing exploration and reshaping the reward landscape?}

\textbf{Formulation.}
To this end, we propose \emph{Find, Fix, Reason} (FFR), a framework that diagnoses missing spatiotemporal dependencies in a student model’s rollouts and, via predefined tools, retrieves minimal task–relevant context from the same video to avoid biased multi–round search. This targeted intervention improves both accuracy and exploration efficiency for small students while leaving the task and policy unchanged. Specifically, let $\pi_{\theta_{\text{old}}}$ be the sampling policy and $\pi_\theta$ the policy to optimize. 
For a video--question pair $x$, sample $G$ first-pass rollouts $\{\tau_i\}_{i=1}^{G}\sim\pi_{\theta_{\text{old}}}$ and score $R(\tau_i)$ each with a verifiable checker $\mathcal{V}$. 
Define the first-pass correctness indicator as follows:
\begin{equation}
z_i \triangleq \mathbb{1}\{\text{correct}(\tau_i)\} \in \{0,1\}.
\end{equation}


\begin{table*}[t]
    \centering
    \small 
    \renewcommand{\arraystretch}{0.85} 
    \caption{Snapshot of teacher's negative constraint strategy across diverse scenarios. The mechanism leverages the teacher model's capability to ensure zero-leakage by converting answers into heuristic instructions.}
    \label{tab:teacher_strategy}
    
    \begin{tabularx}{\textwidth}{@{} l l X @{} }
        \toprule
        \textbf{Scenario} & \textbf{Violation to Avoid (Direct Leakage)} & \textbf{Guided Exploration (FFR Heuristic Instruction)} \\ 
        \midrule
        \textbf{Counting} & ``There are exactly 3 people in frame 15.'' & ``Recount the subjects within the frame range [13, 17].'' \\
        \textbf{Dynamics} & ``The person moves from left to right.'' & ``Track and analyze the subject's movement direction across the provided clip.'' \\
        \textbf{Temporal} & ``The event happens before the person sits down.'' & ``Examine the temporal relationship and causal sequence between the two identified events.'' \\
        \textbf{Spatial} & ``The answer is C / the yellow sphere.'' & ``Identify paths intersected by the static cylinder to determine potential collisions.'' \\
        \textbf{Attribute} & ``The object being picked up is a red metal cube.'' & ``Observe the visual features (color/material) of the object involved in the interaction.'' \\
        \textbf{Logic} & ``The ball stops because it hits the wall.'' & ``Analyze the interaction between the moving object and its surrounding environment.'' \\
        \bottomrule
    \end{tabularx}
    \vspace{-1em}
\end{table*}
\newcommand{\sg}[1]{\operatorname{sg}\!\left[#1\right]}

\textbf{Teacher evidence with no-leakage.}
We introduce a \textbf{frozen} teacher model $\mathcal{T}$ with parameters $\psi$, which generates an
\emph{error type} $e_i \in \mathcal{E}$ and a \emph{minimal evidence patch} $c_i$
based on a student data sample $\mathcal{S}_i$. The teacher policy $\pi_\mathcal{T}$ generates feedback conditioned on this data package, where $\mathcal{S}_i$ may vary depending on the task scenario. Specifically:
\begin{equation}
\label{eq:context_patch}
(e_i, c_i) \sim p_{\mathcal{T}}\!\left( e, c \,\middle|\, \mathcal{S}_i; \sg{\psi} \right),
\qquad
\nabla_{\psi}\, p_{\mathcal{T}} \equiv 0 .
\end{equation}
In typical teaching scenarios, the teacher has access to both the input $x$ and the ground-truth (GT) $y$, and the data package $\mathcal{S}_i = (x, y, \tau_i)$ is used to generate error types $e_i$ and evidence patches $c_i$. In more challenging settings, where the teacher does not have access to the correct answer $y$, the teacher relies on a reduced data package $\mathcal{S}_i = (x, \tau_i)$ to generate error types and evidence based solely on the input and the initial student trajectory. Regardless of the scenario, the evidence $e_i$ and patch $c_i$ provided by the teacher must not directly reveal the correct answer. 

We implement a multi-layered guidance control mechanism to prevent the teacher from providing ``shortcuts''---direct clues that allow the student to bypass the reasoning chain---which would probably otherwise lead to a degradation of the student’s exploration and generalization capabilities. Since traditional metrics are insufficient to capture all edge cases of leakage across diverse video tasks, we leverage the In-Context Learning (ICL) and strong instruction-following abilities of the teacher model itself. 

Specifically, we employ \textbf{negative prompting} to steer the teacher toward identifying \textbf{essential task-related dependencies} (e.g., causal interactions or spatial occlusions) while strictly prohibiting the disclosure of ground-truth options or terminal states. Furthermore, to avoid ``over-guidance'' in the visual domain, the teacher is instructed to provide \textbf{temporal ranges} (e.g., frame intervals) rather than pinpointing a specific ``answer frame.'' This design forces the student model to actively parse the spatiotemporal dynamics within the suggested window to ``re-discover'' the evidence independently. A snapshot of our guidance strategy is summarized in Table~\ref{tab:teacher_strategy}. The detailed definition of error types and prompts for this controlled teacher guidance are provided in the \textbf{Appendix}.

The error taxonomy label $e_i$ and its task-relevant visual evidence $c_i$ if necessary, exposing only task-relevant dependencies and withholding any direct answers. If $z_i=0$, generate a patched solution and evaluate its reward:
\begin{equation}
\tau_i^{*} \sim \pi_{\theta_{\text{old}}}(\,\cdot\,\mid x,c_i),~
R(\tau_i^{*}) \in \{0, 1\} .
\end{equation}

\textbf{Chosen rollout and per-item scalar.}
Define the \emph{chosen} rollout:
\begin{equation}
\widehat{\tau}_i =
\begin{cases}
\tau_i, & z_i=1,\\
\tau_i^{*}, & z_i=0.
\end{cases}
\end{equation}

Use the following scalar for group standardization (``strict repair'' with an optional patch tax $\kappa\ge 0$) and consider the format reward, namely RIR:
\begin{equation}
\label{eq:reward}
\widetilde{R}_i \;=\; z_i(\,R(\tau_i)+R_\text{fmt}(\tau_i)) \;+\; (1-z_i)\,\big(R(\tau_i^{*})+R_\text{fmt}(\tau_i^{*})-\kappa\big).
\end{equation}
Standardize within the $G$ items to obtain the advantage, we can get:
\begin{equation}
A_i \;=\; 
\frac{\widetilde{R}_i \;-\; \operatorname{mean}\!\big(\{\widetilde{R}_j\}_{j=1}^{G}\big)}
{\operatorname{std}\!\big(\{\widetilde{R}_j\}_{j=1}^{G}\big)} .
\end{equation}

\textbf{Final GRPO objective with chosen rollout.}
Let $x_i=(v_i,q_i)$ denote the video–question input, $c_i^\dagger \triangleq c_i$ if $z_i=0$ (patched) and $c_i^\dagger \triangleq \varnothing$ otherwise.
Let the chosen rollout decode an output token sequence 
$y_i^\dagger = (y_{i,1}^\dagger,\ldots,y_{i,T_i}^\dagger)$ with $T_i=|\widehat{\tau}_i|$.
The token-level importance ratio becomes
\begin{equation}
r_{i,t}(\theta) \triangleq 
\frac{\pi_\theta\!\left(y_{i,t}^\dagger \mid v_i, q_i, c_i^\dagger, y_{i,<t}^\dagger\right)}
{\pi_{\theta_{\text{old}}}\!\left(y_{i,t}^\dagger \mid v_i, q_i, c_i^\dagger, y_{i,<t}^\dagger\right)},
\qquad t=1,\ldots,T_i .
\end{equation}
The GRPO objective retains its standard form, replacing the chosen rollout $\widehat{\tau}_i$ and group-standardized advantage $A_i$.
\begin{equation}
\begin{aligned}
\mathcal{J}_{\text{FFR}}(\theta)
&=
\frac{1}{\sum_{i=1}^{G} |\widehat{\tau}_i|}\,
\sum_{i=1}^{G}\sum_{t\in \widehat{\tau}_i}
\mathrm{CLIP}\!\big(r_{i,t}(\theta),\,A_i,\,\epsilon\big)
\\[0.3em]
&\quad
-\;\beta\,\mathbb{D}_{\mathrm{KL}}\!\big[\pi_\theta \,\|\, \pi_{\mathrm{ref}}\big] .
\end{aligned}
\end{equation}
Here $\pi_{\mathrm{ref}}$ is a fixed reference policy and $\epsilon$ is the clipping threshold. Only tokens from the chosen rollout $\widehat{\tau}_i$ contribute to the objective.

\textbf{Discussion.}
Existing methods leverage two-stage training to enable a single model to perform tool invocation or multi-tasking (\eg, captioning, grounding). By setting a callback mechanism, a smaller model is prompted to explore multiple rounds of key context matching, dynamically retrieving finer-grained context. The core insight is access to relevant context rather than tool use itself. Ideally, these methods improve downstream performance by determining when and which tools to call for effective context retrieval. However, the model’s parameter scale presents a challenge: when faced with complex problems, this mechanism often leads the model into repetitive cycles of ``self-doubt''. Specifically, after multiple rounds of tool calls and context retrievals, the model struggles to grasp the key dependencies, resulting in ineffective rollouts and limited on-policy exploration within the framework. Moreover, these pipelines require curated data and costly two-stage tuning.
In contrast, these issues are mitigated by introducing a larger teacher model in FFR framework
. The teacher model, with its superior knowledge base, can accurately identify key dependencies, pinpoint errors in the student’s reasoning, and guide the model by selecting appropriate tools for context retrieval. Specifically, the teacher-generated patch $c_i$ isolates the missing dependency and turns failures into informative observations; typically $R(\tau_i^{*})>R(\tau_i)$, which lifts $\widetilde{R}_i$ and the standardized $A_i$, thereby upweighting evidence-aligned tokens through $\mathrm{CLIP}(r_{i,t},A_i,\epsilon)$ in $\mathcal{J}_{\text{FFR}}(\theta)$.
Because $r_{i,t}(\theta)$ is computed under the same observation $(v_i,q_i,c_i^\dagger)$ for both policies, the update stays strictly on-policy while steering probability mass toward the causal continuation. The framework is plug-and-play—no extra pretraining data—and can be dropped into the training loop of any reasoner.

\begin{table*}[t]
\caption{Performance comparison of video reasoners across benchmarks.
Yellow areas indicate the SFT base models, while the \textbf{bolded} blue cells show results with FFR.
$\Delta\%$ is computed relative to the corresponding SFT baseline (yellow row).}
\label{tab:main_results}
\centering
\resizebox{\textwidth}{!}{%
\renewcommand{\arraystretch}{1.15}
\setlength{\tabcolsep}{4pt}
\begin{tabular}{l cccc cccc}
\toprule
\multirow{2}{*}{\textbf{Model}} 
& \multicolumn{4}{c}{\textbf{Video Reasoning Benchmark}} 
& \multicolumn{4}{c}{\textbf{Video General Benchmark}} \\
\cmidrule(lr){2-5} \cmidrule(lr){6-9}
& \textbf{MMVU$\uparrow$} & \textbf{VSI-Bench$\uparrow$} & \textbf{VideoMMMU$\uparrow$} & \textbf{Video-Holmes$\uparrow$} 
& \textbf{LongVideoBench$\uparrow$} & \textbf{LVBench$\uparrow$} & \textbf{MVBench$\uparrow$} & \textbf{TempCompass$\uparrow$} \\
\midrule
GPT-4o & 75.4 & 34.0 & 61.2 & 42.0 & 58.5 & 48.9 & 64.6 & 73.75 \\
GLM-4.5V  & 68.7 & -- & 72.4 & -- & -- & 53.8 & 73.0 & -- \\
\midrule
Qwen2.5-VL & 59.2 & 27.7 & 47.8 & 27.8 & 43.2 & 31.6 & 57.4 & 72.2 \\
Video-ChatR1 & 64.2 & -- & 48.9 & 35.7 & 49.1 & 34.3 & 67.9 & 72.9 \\ 
AVATAR & 65.6 & -- & -- & 45.1 & -- & 38.4 & 66.4 & -- \\
Pixel-Reasoner & 64.0 & 30.2 & 49.3 & 35.1 & 52.6 & 34.3 & 67.8 & 75.4 \\
Video-Thinker & 62.4 & 26.3 & 43.9 & 43.2 & 48.3 & 37.0 & 58.9 & 67.5 \\
\midrule
\rowcolor{yellow!10}
Video-R1-SFT & 61.3 & 31.8 & 47.4 & 34.6 & 47.6 & 30.7 & 59.4 & 69.2 \\
Video-R1 & 63.8 & 35.8 & 52.3 & 36.5 & 52.7 & 35.3 & 63.9 & 73.2 \\
\rowcolor{cyan!10}
$+$ FFR & \textbf{68.5} & \textbf{38.9} & \textbf{54.6} & \textbf{52.3} & \textbf{55.3} & \textbf{38.1} & \textbf{68.8} & \textbf{75.6} \\
\rowcolor{cyan!10}
$\Delta\%$ & \textcolor{blue}{+11.75} & \textcolor{blue}{+22.33} & \textcolor{blue}{+15.19} & \textcolor{blue}{+51.16} & \textcolor{blue}{+16.18} & \textcolor{blue}{+24.10} & \textcolor{blue}{+15.82} & \textcolor{blue}{+9.25} \\
\rowcolor{yellow!10}
VideoRFT-SFT & 60.5 & 31.7 & 48.5 & 27.1 & 47.3 & 26.9 & 57.0 & 68.4 \\
VideoRFT & 68.5 & 36.8 & 51.1 & 40.0 & 52.5 & 33.9 & 62.1 & 73.7 \\
\rowcolor{cyan!10}
$+$ FFR & \textbf{70.1} & \textbf{38.6} & \textbf{54.9} & \textbf{48.0} & \textbf{54.9} & \textbf{37.8} & \textbf{68.2} & \textbf{75.4} \\
\rowcolor{cyan!10}
$\Delta\%$ & \textcolor{blue}{+15.87} & \textcolor{blue}{+21.77} & \textcolor{blue}{+13.20} & \textcolor{blue}{+77.12} & \textcolor{blue}{+16.07} & \textcolor{blue}{+40.52} & \textcolor{blue}{+19.65} & \textcolor{blue}{+10.23} \\
\bottomrule
\end{tabular}
}
\end{table*}

\begin{table*}[t]
\caption{Ablation study of the FFR framework on video reasoning and general video understanding benchmarks.}
\label{tab:ablation_FFR}
\centering
\resizebox{\textwidth}{!}{%
\renewcommand{\arraystretch}{1.15}
\setlength{\tabcolsep}{4pt}
\begin{tabular}{l cccc cccc}
\toprule
\multirow{2}{*}{\textbf{Model}} & \multicolumn{4}{c}{\textbf{Video Reasoning Benchmark}} & \multicolumn{4}{c}{\textbf{Video General Benchmark}} \\
\cmidrule(lr){2-5} \cmidrule(lr){6-9}
& \textbf{MMVU$\uparrow$} & \textbf{VSI-Bench$\uparrow$} & \textbf{VideoMMMU$\uparrow$} & \textbf{Video-Holmes$\uparrow$} & \textbf{LongVideoBench$\uparrow$} & \textbf{LVBench$\uparrow$} & \textbf{MVBench$\uparrow$} & \textbf{TempCompass$\uparrow$} \\
\midrule
Qwen2.5-VL & 59.2 & 27.7 & 47.8 & 27.8 & 43.2 & 31.6 & 57.4 & 72.2 \\
w./ SFT & 61.3 & 31.8 & 47.4 & 34.6 & 47.6 & 30.7 & 59.4 & 69.2 \\
Only w./ vanilla GRPO & 60.3 & 28.4 & 42.9 & 45.6 & 51.9 & 33.9 & 60.0 & 69.7 \\
\rowcolor{yellow!10}
w./ SFT + w./ vanilla GRPO & 61.3 & 32.8 & 43.4 & 46.2 & 51.2 & 36.7 & 61.0 & 70.5 \\
\rowcolor{yellow!10}
w./ SFT + w./ T-GRPO (Video-R1) & 63.8 & 35.8 & 52.3 & 36.5 & 52.7 & 35.3 & 63.9 & 73.2 \\
\rowcolor{yellow!10}
w./ FFR (no visual context) & 64.4 & 36.6 & 54.0 & 42.3 & 51.6 & 36.7 & 60.8 & 73.4 \\
\rowcolor{yellow!10}
w./ FFR (no GT reference) & 63.7 & 38.4 & 51.3 & 44.7 & 54.4 & 38.0 & 62.6 & 75.4 \\
\rowcolor{cyan!10}
Full model & \textbf{68.5} & \textbf{38.9} & \textbf{54.6} & \textbf{52.3} & \textbf{55.3} & \textbf{38.1} & \textbf{68.8} & \textbf{75.6} \\
\bottomrule
\end{tabular}
}
\end{table*}

\section{Experiment}
\label{sec:experiment}
\subsection{Experiment Setting}

\textbf{Benchmarks.}
We evaluate on four \emph{video reasoning} benchmarks—MMVU~\citep{MMVU}, VSI-Bench~\citep{VSIBench}, VideoMMMU~\citep{VideoMMMU}, and Video-Holmes~\citep{cheng2025video}—covering expert-level understanding, spatial reasoning, domain knowledge, and causal-narrative inference, respectively. We further include four \emph{general video understanding} benchmarks—LongVideo-Bench~\citep{wu2024longvideobench}, LVBench~\citep{wang2025lvbench}, MVBench~\citep{MVBench}, and TempCompass~\citep{TempCompass}—to assess generalization across long-form comprehension, temporal consistency, and motion-based reasoning. Detailed descriptions are in the \textbf{Appendix}.

\textbf{Baselines.}
We compare FFR against on-policy methods (Qwen2.5-VL-7B~\citep{bai2025qwen2}, Video-R1-7B~\citep{VideoR1}, Video-RFT-7B~\citep{VideoRFT}, Video-ChatR1-7B~\citep{VideoChatR1}), mixed-policy methods (AVATAR-7B~\citep{kulkarni2025avatar}), and tool-use reasoners (Pixel-Reasoner~\citep{su2025pixel}, Video-Thinker~\citep{wang2025video}), as illustrated in Figure~\textcolor{magenta}{\ref{fig:comp}}.

\textbf{Implementation Details.}
We use Video-R1-7B-SFT and VideoRFT-7B-SFT as base models, training on 8$\times$A100-80G GPUs. Video inputs are limited to 16 frames at 128$\times$28$\times$28 resolution. The training dataset contains 4,000 samples sourced from Video-R1, Video-Thinker, and Video-Holmes, trained for 1 epoch with learning rate 5e-6 and 8 rollouts per sample. We select GLM-4.5V as the teacher model based on cost-effectiveness analysis (Figure~\textcolor{magenta}{\ref{fig:analysis}}). The framework is built on R1-V~\citep{chen2025r1v}. More details are in the \textbf{Appendix}.

\subsection{Main Results}
\textbf{How well can FFR Perform?}
FFR significantly outperforms traditional RFT methods, as demonstrated in Table~\ref{tab:main_results}, with substantial improvements across both video reasoning and general benchmarks. For instance, using Video-R1-SFT as the base model, FFR leads to impressive gains in VSI-Bench (+22.33\%) and MMVU (+11.75\%), and a similar performance boost is observed with VideoRFT-SFT, where FFR improves VSI-Bench (+21.77\%) and MMVU (+15.87\%), both showing significant advantages over other RFT methods. This demonstrates FFR's ability to consistently enhance model performance, regardless of the base model.
In addition to improving performance, FFR also enables smaller student models to match or even surpass the performance of larger teacher models like GLM-4.5V, showcasing its potential for model compression. Specifically, with Video-R1-SFT as the base model, FFR-enhanced models achieve comparable performance to GLM-4.5V on MMVU, and when using VideoRFT-SFT as the base model, FFR further boosts performance, exceeding the results of GLM-4.5V. This highlights FFR's ability to compress models without sacrificing accuracy, providing an efficient solution for deploying high-performance models.
Moreover, when compared to replay-based methods (e.g., AVATAR) and tool-use methods (e.g., VideoThinker and PixelReasoner), FFR consistently delivers superior performance by leveraging enhanced repair contexts from the teacher model, facilitating more effective exploration. This is particularly evident in complex video reasoning tasks such as Video-Holmes and VSI-Bench, where FFR results in substantial improvements, underscoring its effectiveness in tackling challenging video understanding problems.

\begin{table}[t]
\centering
\small
\caption{SFT-Teacher vs.\ FFR on video reasoning benchmarks. FFR outperforms SFT at every teacher scale. Full results in \textbf{Appendix} (Table~\ref{tab:sft_comparison_full}).}
\label{tab:sft_comparison}
\setlength{\tabcolsep}{3.5pt}
\renewcommand{\arraystretch}{1.1}
\begin{tabular}{l cccc c}
\toprule
\textbf{Method} & \textbf{MMVU} & \textbf{VSI} & \textbf{VMMU} & \textbf{V-Holm.} & \textbf{Avg.} \\
\midrule
Video-R1-SFT & 61.3 & 31.8 & 47.4 & 34.6 & 43.8 \\
\midrule
SFT (32B) & 63.9 & 39.1 & 42.0 & 43.3 & 47.1 \\
SFT (235B) & 67.4 & \textbf{41.9} & 46.2 & 47.1 & 50.7 \\
\midrule
\rowcolor{cyan!10}
FFR (32B) & 67.9 & 38.5 & 50.5 & 47.8 & 51.2 \\
\rowcolor{cyan!10}
FFR (235B) & 68.2 & 38.1 & \textbf{56.5} & 51.6 & 53.6 \\
\rowcolor{cyan!10}
FFR (GLM-4.5V) & \textbf{68.5} & 38.9 & 54.6 & \textbf{52.3} & \textbf{53.6} \\
\bottomrule
\end{tabular}
\vspace{-1em}
\end{table}

\subsection{Ablation Studies}
\textbf{What actually matters in FFR?} Table~\ref{tab:ablation_FFR} isolates each component. We clarify two key ablation configurations: \emph{no visual context} means the evidence patch $c_i$ contains only textual guidance (error classification, temporal markers) while omitting visual cues such as key-frame indices and spatial regions; \emph{no GT reference} means the teacher operates with reduced input $\mathcal{S}_i = (x, \tau_i)$ in Eq.~\ref{eq:context_patch}, diagnosing errors without access to the ground-truth answer $y$.
Three insights emerge:
\ding{182}~The teacher mechanism is the primary performance driver---removing it entirely (vanilla GRPO) yields the weakest overall reasoning performance. Notably, vanilla GRPO scores higher than T-GRPO (Video-R1) on Video-Holmes; we attribute this to T-GRPO's task-specific reward shaping, which narrows the policy at the cost of generalization to Video-Holmes's causal-narrative format.
\ding{183}~Visual context is critical for spatiotemporal tasks: the full model outperforms the no-visual-context variant by $+$10.0 on Video-Holmes, confirming that key-frame and spatial evidence are essential for complex reasoning.
\ding{184}~GT reference improves patch precision: with access to $y$, the teacher generates more targeted patches, yielding $+$7.6 on Video-Holmes over the no-GT variant.

\textbf{Why Not Directly Distill Teacher Traces?}
A conventional alternative is Supervised Fine-Tuning (SFT) on teacher-generated reasoning traces. However, as shown in Table~\ref{tab:sft_comparison}, FFR consistently outperforms SFT across all teacher scales; notably, FFR with a 32B teacher already surpasses SFT with a 235B model (51.2 vs. 50.7). This performance gap underscores that FFR's efficacy stems from its targeted intervention mechanism rather than mere exposure to high-quality data. While SFT forces a wholesale imitation of the teacher's output distribution, FFR acts as a heuristic guide: it intervenes only on failed rollouts to expose the underlying spatiotemporal dependencies. By internalizing these critical reasoning anchors---rather than just mimicking surface-level traces---the student learns to reconstruct robust causal paths while maintaining the benefits of on-policy exploration.

\begin{figure*}[t]
\centering
\includegraphics[width=\textwidth]{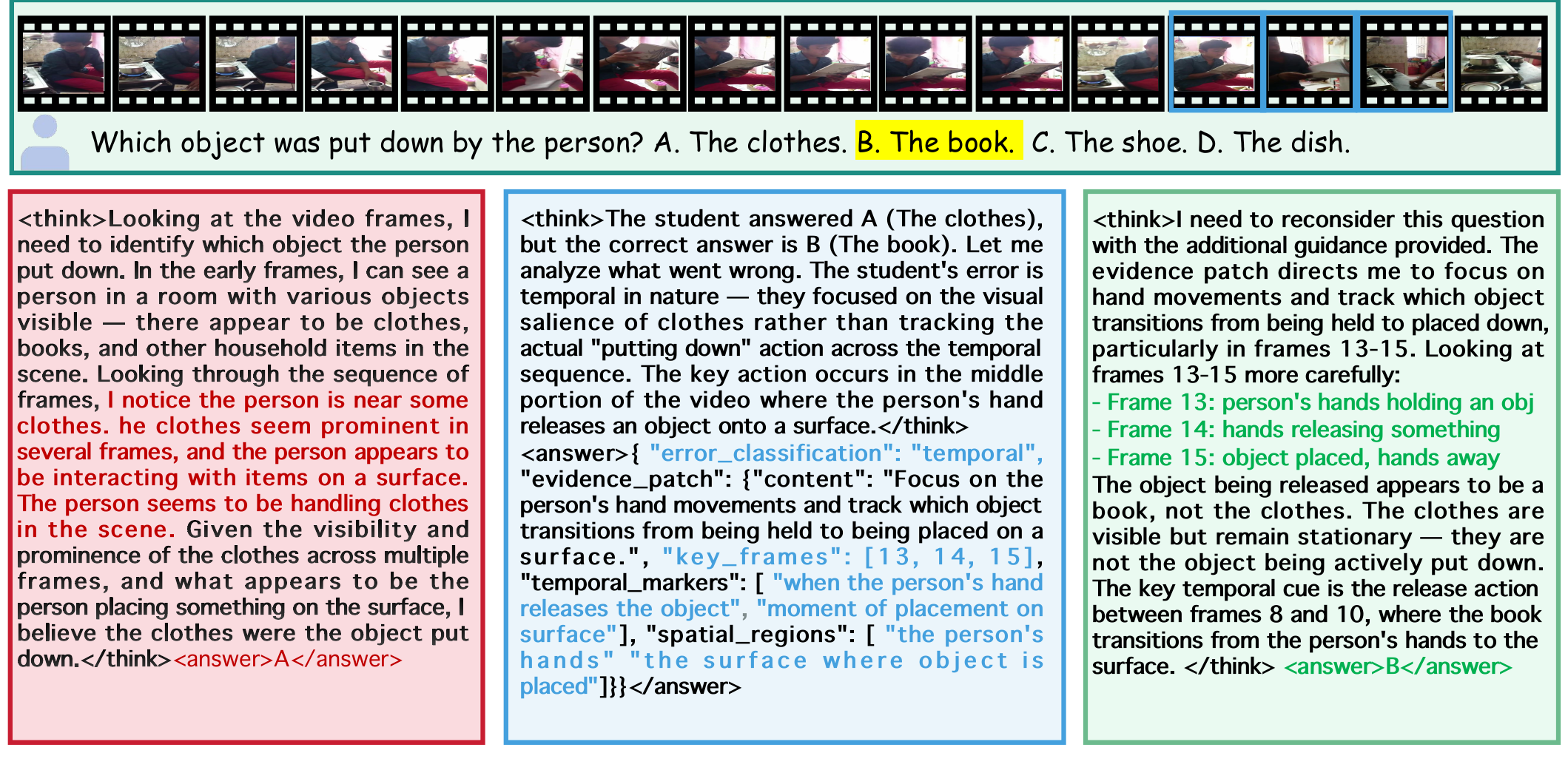}
\vspace{-1em}
\caption{Case study of FFR's intervention on a STAR training sample. \textbf{Top:} 16 uniformly sampled video frames with key frames 13--15 highlighted; the question asks which object was put down. \textbf{Bottom left:} the student's first rollout incorrectly answers A (clothes). \textbf{Bottom center:} the teacher diagnoses a \textit{temporal\_error} and generates an evidence patch pointing to key frames and temporal markers. \textbf{Bottom right:} guided by the patch, the student re-examines frames 13--15 and correctly answers B (book). The patch redirects attention without revealing the answer.}
\label{fig:case_study}
\end{figure*}

\begin{figure}[t]
    \centering
    \includegraphics[width=\columnwidth]{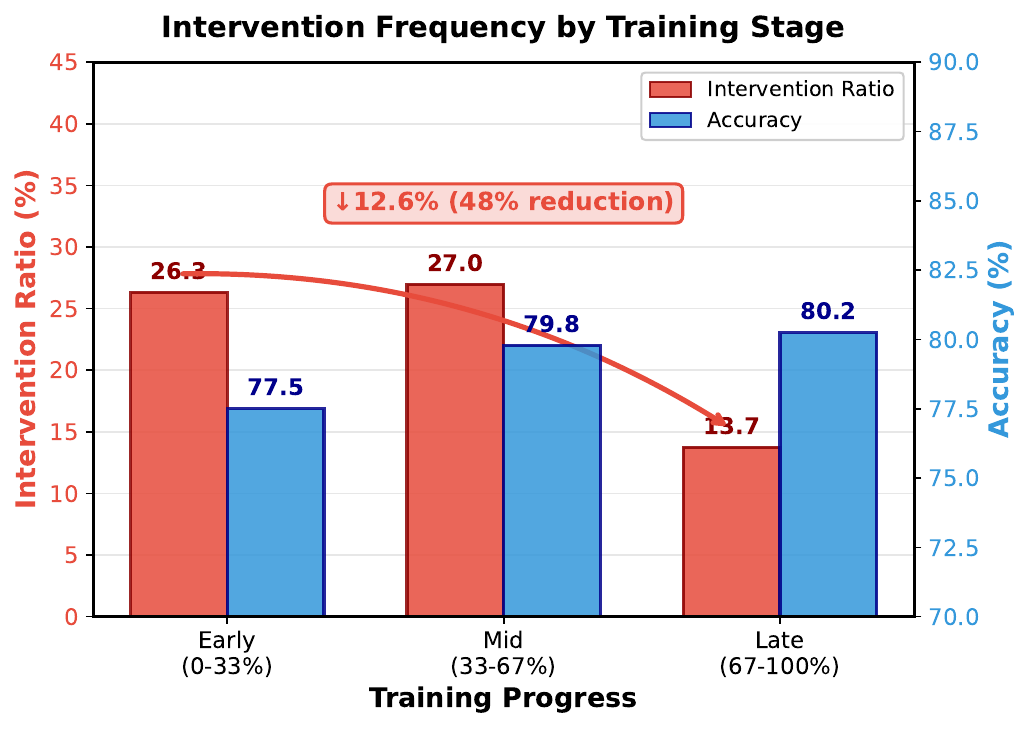}
    \caption{Intervention ratio and accuracy across training stages. The intervention ratio drops 48\% from early to late training while accuracy improves, indicating capability internalization.}
    \label{fig:intervention}
    \vspace{-1.2em}
\end{figure}

\begin{table}[t]
\centering
\small
\caption{Answer leakage rates under different prompt configurations (200 manually verified interactions).}
\label{tab:leakage_validation}
\setlength{\tabcolsep}{4pt}
\renewcommand{\arraystretch}{1.1}
\begin{tabular}{lccc}
\toprule
\textbf{Prompt Config.} & \textbf{Direct} & \textbf{Partial} & \textbf{Total} \\
\midrule
No constraints & 15.5\% & 24.0\% & 39.5\% \\
+ Format constraints & 6.0\% & 13.5\% & 19.5\% \\
+ Negative prompt & 1.5\% & 4.0\% & 5.5\% \\
+ Both (final) & 0\% & 0\% & 0\% \\
\bottomrule
\end{tabular}
\vspace{-1em}
\end{table}

\subsection{Teacher Guidance Validation}
A central question is whether FFR's improvements arise from its specific intervention mechanism or merely from exposure to higher-quality trajectories. We provide mechanism-focused analyses below.

\textbf{Case Study: Evidence Patch $\rightarrow$ Corrected Reasoning.}
Figure~\ref{fig:case_study} traces the full FFR pipeline on a real training sample: the student misidentifies the put-down object (Answer~A), the teacher diagnoses a \textit{temporal\_error} and provides an evidence patch containing key frames [13--15] and temporal markers (``when hand releases object''). The patch does \emph{not} reveal the answer---it redirects attention so the student must re-examine the specified frames and independently identify the correct object (Answer~B). This exemplifies FFR's design principle: \emph{guiding exploration without bypassing reasoning}.

\textbf{Answer Leakage Prevention.}
A concern is whether evidence patches inadvertently reveal the answer. Table~\ref{tab:leakage_validation} validates our prompt design: combining structured output constraints with explicit negative prompts reduces leakage to 0\% across 200 manually verified interactions (see \textbf{Appendix} for complete prompt designs). While the sample size bounds the 95\% CI upper limit at ${\sim}$1.5\%, we further note that even hypothetical leakage would effectively reduce FFR to an SFT-like paradigm---the student would receive correct trajectories ``for free.'' Since SFT-Teacher already performs substantially below FFR (Table~\ref{tab:sft_comparison}), leakage alone cannot explain the observed gains.

\textbf{Intervention Dynamics Analysis.}
Figure~\ref{fig:intervention} shows the intervention ratio drops from 26.3\% in the early stage to 13.7\% by the late stage---a \textbf{48\% relative reduction}---while training accuracy improves from 77.5\% to 80.2\%. This divergence indicates capability internalization: the student successfully substitutes external evidence patches with autonomous spatiotemporal reasoning. Rather than becoming dependent on the teacher, the student leverages heuristic instructions (Table~\ref{tab:teacher_strategy}) to bridge reasoning gaps without answer leakage, confirming that FFR enables \textbf{guided exploration} rather than serving as a training crutch.

\begin{figure}[t]
    \centering
    \includegraphics[width=\linewidth]{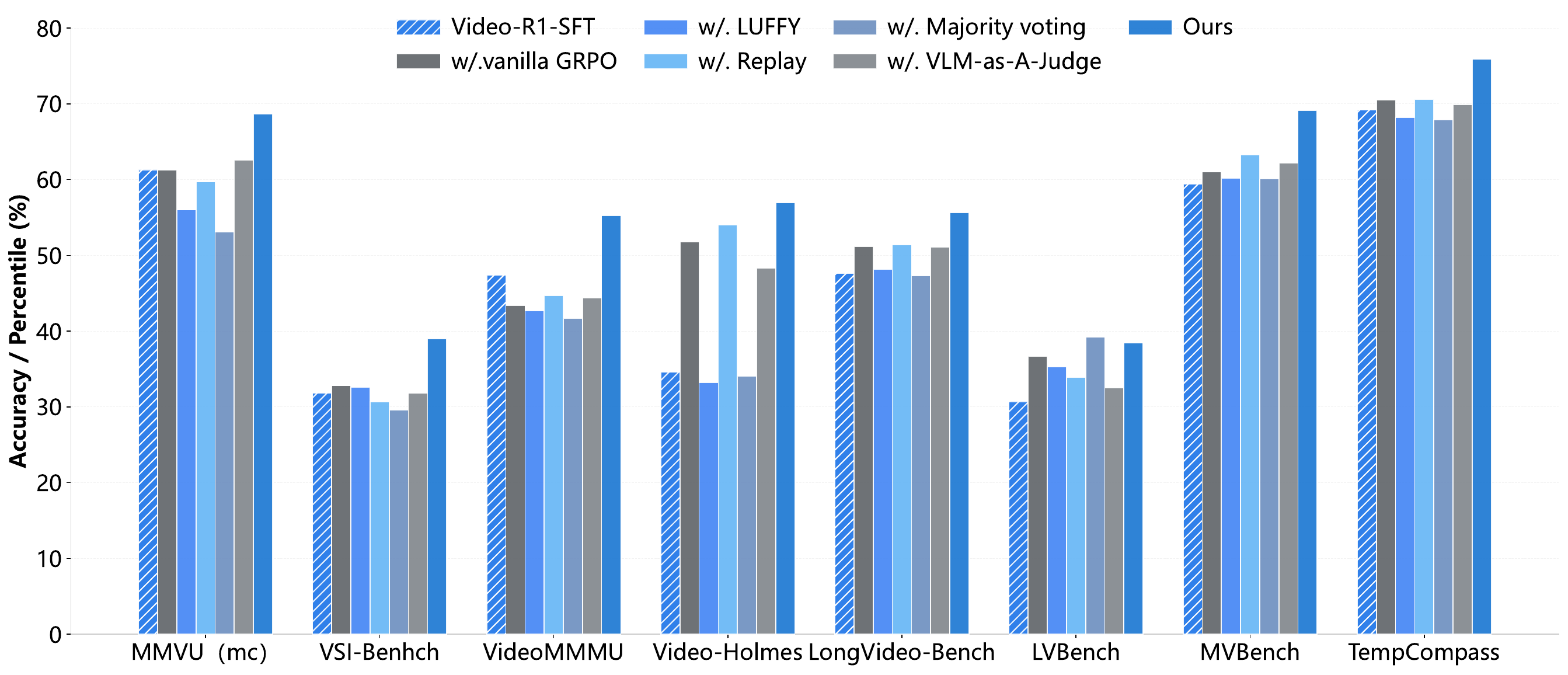}
    \vspace{-2em}
\caption{Comparison with other RL methods in Video Reasoning.}
    \label{fig:chart}
    \vspace{-2em}
\end{figure}

\subsection{Further Analyses}

\textbf{Evaluation of More RL Methods in Video Reasoning.}  
To compare the effectiveness of prevalent RL methods in video reasoning, we selected hybrid policy approaches from \citep{yan2025learning} (LUFFY) and \citep{zhan2025exgrpo} (Replay), as shown in Figure~\textcolor{magenta}{\ref{fig:comp}}(\textit{b}), and validated them following their experimental setups. Inspired by \citep{zuo2025ttrl}, we incorporated a multi-round voting mechanism to assess its performance against labeled data. Since the current binary accuracy reward offers limited protection against reward-hacking, we also explored replacing it with a continuous 0–1 reward using VLMs as judges. These methods, along with our baseline (Video-R1-SFT, vanilla GRPO) and the proposed FFR, were validated on eight datasets, with results shown in Figure~\textcolor{magenta}{\ref{fig:chart}}. Three takeaways can be drawn: \ding{182} The majority voting mechanism does not show significant improvement when labeled data is available;  \ding{183} Replay-based methods, which replay low-entropy, stable reasoning trajectories of similar difficulty, effectively enhance performance during inference; \ding{184} Under the off-policy guidance condition, directly using the reward provided by the off-policy method (VLM-as-A-Judge) outperforms relying on the reasoning trajectory itself (LUFFY). However, dynamically acquiring more accurate context leads to the most substantial improvements for the on-policy model, as evidenced by the superior performance of the proposed FFR across all benchmarks.

\begin{figure}[t]
    \centering
    \begin{subfigure}[b]{0.55\columnwidth}
        \centering
        \includegraphics[width=\textwidth]{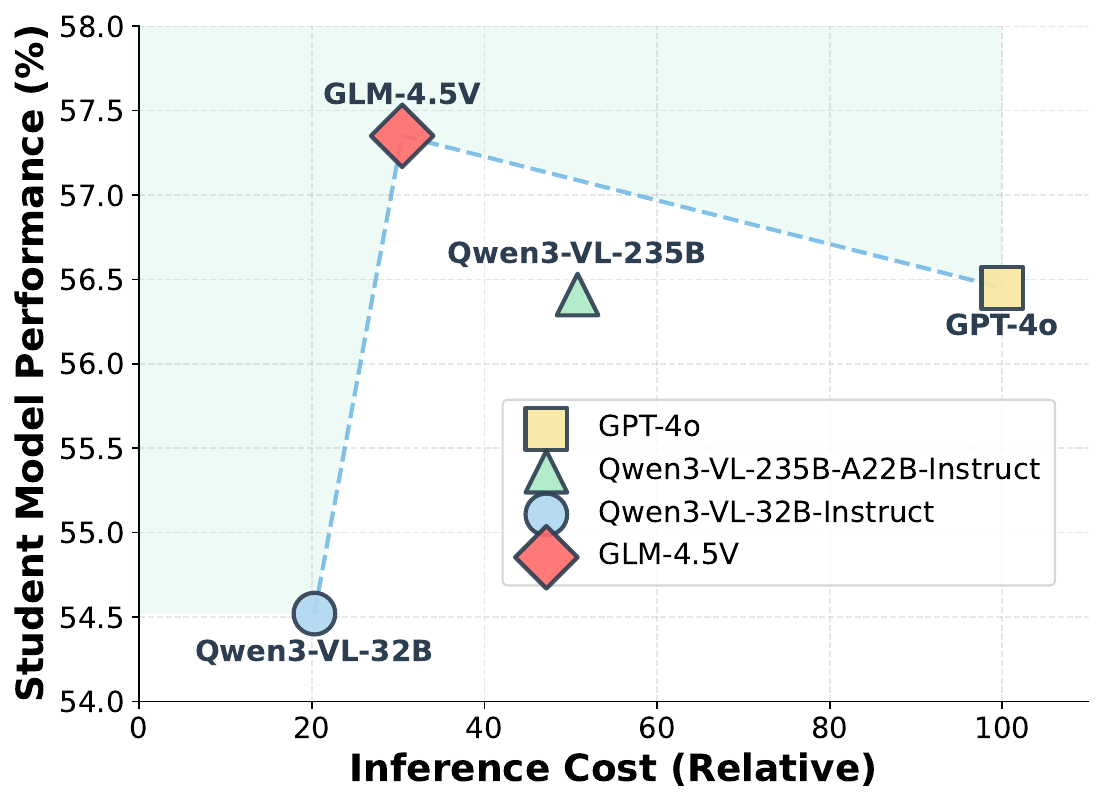}
        \vspace{-1.4em}
        \caption{Pareto Analysis}
        \label{fig:pareto}
    \end{subfigure}
    \hfill
    \begin{subfigure}[b]{0.42\columnwidth}
        \centering
        \includegraphics[width=\textwidth]{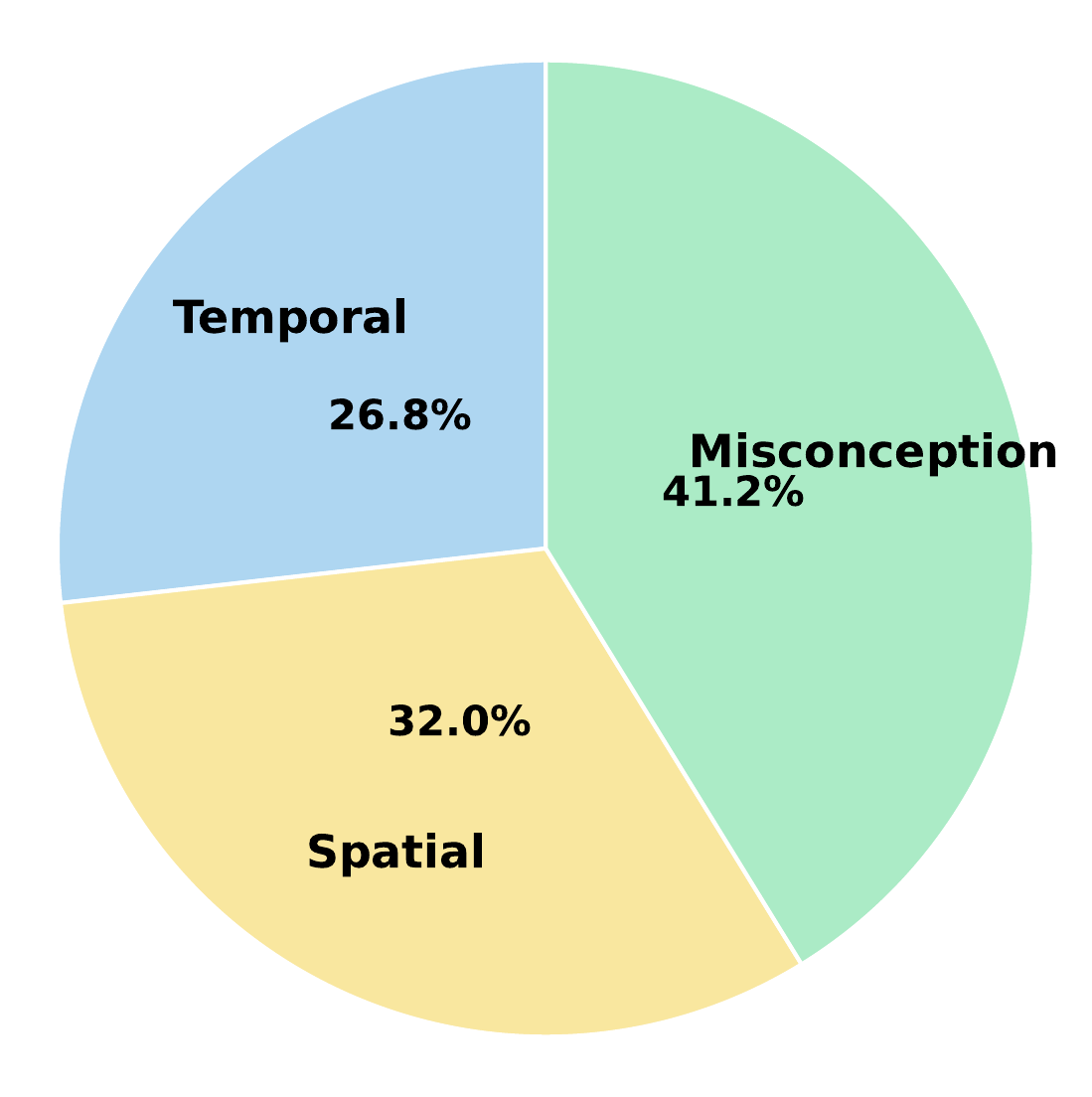}
                \vspace{-1.4em}
        \caption{Error Types}
        \label{fig:errors}
    \end{subfigure}
    \caption{Teacher model selection and error analysis.}
    \label{fig:analysis}
    \vspace{-1.5em}
\end{figure}

\textbf{Teacher Model Selection and Error Distribution.}
Figure~\textcolor{magenta}{\ref{fig:analysis}}(\textit{a}) shows GLM-4.5V achieves Pareto optimality in the accuracy-cost trade-off among four candidate teachers (details in \textbf{Appendix}). Figure~\textcolor{magenta}{\ref{fig:analysis}}(\textit{b}) reveals that misconception errors (41.2\%) dominate over spatial (32\%) and temporal (26.8\%) errors, indicating students primarily fail due to query misinterpretation rather than visual perception limitations---validating our intervention strategy that targets task-interpretation correction.

\begin{figure}[t]
    \centering
    \includegraphics[width=0.98\columnwidth]{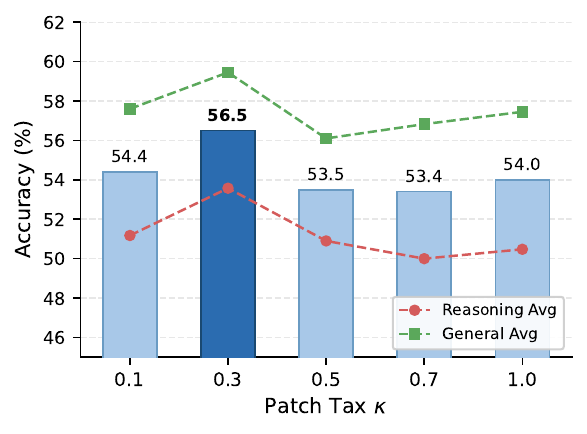}
    \vspace{-1em}
    \caption{Sensitivity to patch tax $\kappa$. Bars show overall average; lines show reasoning vs.\ general benchmark averages. $\kappa{=}0.3$ achieves the best balance. Per-benchmark results in \textbf{Appendix} Table~\ref{tab:kappa_sensitivity}.}
    \label{fig:kappa}
    \vspace{-1.5em}
\end{figure}

\textbf{Sensitivity to Patch Tax $\kappa$.}
The patch tax $\kappa$ in Eq.~\ref{eq:reward} penalizes rollouts that rely on teacher patches, balancing guidance utilization against autonomous reasoning. Figure~\ref{fig:kappa} reports performance across $\kappa \in \{0.1, 0.3, 0.5, 0.7, 1.0\}$. The optimal value is $\kappa{=}0.3$ (56.5\% average): too small a penalty ($\kappa{=}0.1$) insufficiently discourages patch reliance, yielding strong perception but degraded reasoning; conversely, $\kappa \geq 0.5$ excessively discounts teacher contributions. The sweet spot at $\kappa{=}0.3$ enables the student to benefit from teacher guidance while being sufficiently incentivized to internalize reasoning independently.

\textbf{Generalization Across Scales and Architectures.}
A key concern is whether FFR generalizes beyond the 7B Qwen2.5-VL students. We evaluate two axes: \emph{model scale} (Qwen2.5-VL-3B) and \emph{architecture family} (Qwen3-VL-8B). At the 3B scale, FFR boosts the overall average from 34.6 to 45.7 ($+$11.1). On Qwen3-VL-8B (base avg.\ 62.5), FFR lifts the average to 65.1, substantially outperforming vanilla GRPO (62.8); notably, the FFR-trained 8B student surpasses its 32B teacher on Video-Holmes (48.7 vs.\ 42.6) and LongVideoBench (68.4 vs.\ 61.5). This is consistent with FFR's role: the teacher excels at \emph{diagnosing} errors and localizing evidence, a skill that does not require strong end-task accuracy; the student then leverages this diagnostic signal through on-policy exploration to develop more robust answer-generation strategies than the teacher itself. Full per-benchmark results are in the \textbf{Appendix} (Tables~\ref{tab:3b_results},~\ref{tab:qwen3_results}).

\textbf{Inference-Time Frame Scaling.}\enlargethispage{2\baselineskip}
Our model is trained with 16 frames, but at inference time we can increase the frame budget \emph{without retraining}. Scaling from 16 to 64 frames improves the overall average from 56.5 to 57.7, with long-video benchmarks benefiting most (LongVideoBench $+$3.1, LVBench $+$3.9). Short-video benchmarks such as MVBench exhibit non-monotonic behavior (68.8$\rightarrow$61.3$\rightarrow$64.6), likely because redundant frames introduce noise for action-centric questions designed around short clips; nevertheless, 6 of 8 benchmarks improve at 64 frames (full results in \textbf{Appendix} Table~\ref{tab:frame_scaling}).

\section{Conclusion}
\label{sec:conclusion}
We present FFR, a lightweight framework for improving on-policy video reasoning. To overcome sparse rewards and weak spatiotemporal credit assignment, FFR uses a frozen teacher for minimal, non-leaky guidance, steering policy learning to valid visual dependencies while retaining on-policy RLVR optimization. Experiments confirm strong gains on complex long-horizon and fine-grained reasoning tasks. Looking forward, we will improve teacher modeling of visual dependencies and explore integrating FFR with self-distillation, as its refined signals enable self-distillation to explicitly provide accurate token-level credit assignment and stable magnitude-aware learning.

\clearpage
\bibliography{main}
\bibliographystyle{icml2026}

\onecolumn
\appendix
\lstdefinestyle{promptstyle}{
    backgroundcolor=\color{gray!5},
    basicstyle=\ttfamily\scriptsize,  
    breaklines=true,
    breakatwhitespace=false,
    frame=single,
    numbers=none,  
    showstringspaces=false,
    tabsize=2,
    xleftmargin=0.5em,
    xrightmargin=0.5em,
    aboveskip=0.5em,
    belowskip=0.5em,
    columns=flexible,
    keepspaces=true
}

\section{Teacher Model Prompts}
\label{appendix:prompts}

This section provides the complete prompts used for the teacher model in our experiments. These prompts were carefully designed through extensive preliminary testing to ensure optimal compatibility with the teacher model's capabilities, minimal tool invocation errors, and consistent formatted output generation. The fixed prompts presented here represent the best-performing versions selected from multiple iterations based on empirical validation.

\subsection{Teacher Error Analysis Prompt}
\label{appendix:teacher_error}

The \texttt{TEACHER\_ERROR\_ANALYSIS\_PROMPT} is specifically designed for analyzing incorrect responses from the student model while maintaining strict no-leakage constraints. As described in Section~\ref{sec:methodology}, the teacher provides guidance regardless of whether it has access to the ground truth answer, ensuring that the evidence patch never directly reveals the correct solution.

\begin{lstlisting}[style=promptstyle, caption={TEACHER\_ERROR\_ANALYSIS\_PROMPT}]
You are an expert teacher model analyzing incorrect responses from a student 
model's video reasoning.

{input_information}

## Instructions

The student's answer is incorrect. Your task is to:

1. First, think through your analysis in <think> tags:
   - Identify what the student misunderstood
   - Determine the specific type of error
   - Identify minimal evidence that would correct the error

2. Then provide structured output in <answer> tags.

## Error Categories (choose one):

- `temporal`: Misunderstood temporal sequences or event ordering
- `spatial`: Incorrect interpretation of specific frame(s)  
- `misconception`: Misinterpretation of task requirements or question intent

## Output Format

<think>
[Your analysis of the student's error]
</think>

<answer>
{
    "error_classification": "one of: temporal, spatial, misconception",
    "evidence_patch": {
        "content": "Minimal guidance to correct error WITHOUT revealing answer",
        "key_frames": [list of important frame indices],
        "temporal_markers": ["key timestamps or temporal relationships"],
        "spatial_regions": ["important regions in specific frames"]
    }
}
</answer>

## Critical Constraints:
- NEVER directly reveal the correct answer in the evidence patch
- Provide MINIMAL necessary information to guide correction
- Focus on highlighting what was missed, not stating the answer
- The student must still reason independently to reach the solution

## Leakage Prevention for Directly-Observable Queries:
Some queries have answers that are directly observable (e.g., color, count,
object presence/absence, or a single discriminative frame). For these cases,
apply STRICTER constraints:

- Color/Appearance: Do NOT mention the target color or visual attribute.
  Instead: "Re-examine the object's appearance in frame X."
- Object Presence/Absence: Do NOT confirm or deny existence.
  Instead: "Look more carefully at the specified region across frames X-Y."
- Counting: Do NOT state the count.
  Instead: "Recount the items in the specified area, frame by frame."
- Single Discriminative Frame: Do NOT describe the frame content.
  Instead: "Pay closer attention to the transition around frame X."
- Action Identity: Do NOT name the action.
  Instead: "Track the subject's movement trajectory between frames X and Y."

If the answer can be inferred from ANY single field of the evidence patch
alone (key_frames + temporal_markers + spatial_regions + content), the patch
is TOO revealing. Reduce specificity until the student must combine the
patch with their own visual re-examination to arrive at the answer.

Remember: Your evidence must help the student identify their error without
leaking the answer. Guide discovery, don't provide solutions.
\end{lstlisting}

\subsection{Teacher Negative Prompt}
\label{appendix:negative_prompt}

A critical challenge in designing the teacher model is ensuring it provides meaningful guidance without revealing the correct answer. Our preliminary experiments revealed that even state-of-the-art language models tend to inadvertently leak answer information when attempting to help correct errors. To address this, we developed a comprehensive negative prompt that explicitly constrains the teacher's behavior through clear prohibitions and concrete examples.

The \texttt{TEACHER\_NEGATIVE\_PROMPT} serves as a crucial safeguard against answer leakage by establishing strict boundaries on what information the teacher can provide. Rather than directly stating what occurs in the video or providing specific details that would shortcut the student's reasoning process, the teacher is instructed to guide the student toward discovering errors independently. This approach ensures that the learning signal remains meaningful - the student must still perform visual analysis and reasoning to arrive at the correct answer, with the teacher merely highlighting where attention should be focused.

\begin{lstlisting}[style=promptstyle, caption={TEACHER\_NEGATIVE\_PROMPT}]
## Critical Constraints - DO NOT VIOLATE

You MUST NOT:
- State or imply the correct answer directly
- Describe what happens in the video beyond what the student already observed
- Provide specific counts, colors, or object identities unless already mentioned by student
- Complete the student's reasoning for them
- Use phrases like "the answer is", "you should select", or "the correct choice"

## Examples of Violations to Avoid

**Counting Dynamics**
* **BAD:** "There are exactly 3 people in frame 15." (Directly reveals the count)
* **GOOD:** "Recount the subjects within the frame range [13, 17]."

**Dynamics**
* **BAD:** "The person moves from left to right." (Directly describes the action/direction)
* **GOOD:** "Track and analyze the subject's movement direction across the provided clip."

**Temporal**
* **BAD:** "The event happens before the person sits down." (Directly reveals temporal order/causality)
* **GOOD:** "Examine the temporal relationship and causal sequence between the two identified events."

**Spatial**
* **BAD:** "The answer is C / the yellow sphere." (Directly points to the answer/object)
* **GOOD:** "Identify paths intersected by the static cylinder to determine potential collisions."

**Attribute**
* **BAD:** "The object being picked up is a red metal cube." (Directly reveals visual properties)
* **GOOD:** "Observe the visual features (color/material) of the object involved in the interaction."

**Logic**
* **BAD:** "The ball stops because it hits the wall." (Directly reveals the causal explanation)
* **GOOD:** "Analyze the interaction between the moving object and its surrounding environment."

## Filtering Rules for Directly-Observable Answers

When the question asks about a directly observable attribute:
1. NEVER name the attribute value (color, count, object identity, direction).
2. NEVER describe frame content that would make the answer self-evident.
3. ALWAYS redirect to a region or temporal window, requiring the student
   to RE-OBSERVE the visual evidence independently.
4. If unsure whether guidance leaks the answer, apply the "blind test":
   Could someone who has NOT seen the video determine the answer from
   your patch alone? If yes, the patch is too revealing.

Remember: Your role is to help students discover their errors through guided
exploration, not to provide answers. Every piece of evidence should require
the student to observe, analyze, and conclude independently.
\end{lstlisting}

This negative prompt component is integrated with the main teacher prompt to ensure consistent behavior across all interactions. The examples provided illustrate the distinction between revealing information (BAD) and guiding discovery (GOOD), helping the teacher model understand the appropriate level of assistance.

\textbf{Handling Directly-Observable Queries.}
A particular challenge arises when the answer is directly observable---\eg, a color, an object's presence or absence, or a count visible in a single frame. In such cases, even minimal content guidance risks inadvertently revealing the answer. Our prompt design addresses this through three layers: \ding{182}~The error analysis prompt includes \emph{query-type-specific} constraints (see Listing~1) that explicitly prohibit naming colors, counts, or attribute values; \ding{183}~The negative prompt enforces a ``blind test'' rule: if the patch alone (without seeing the video) would allow someone to infer the answer, it must be made less specific; \ding{184}~The structured output format forces the teacher to decompose guidance into \texttt{key\_frames}, \texttt{temporal\_markers}, and \texttt{spatial\_regions}---none of which individually encode the answer. Together, these constraints ensure that the student must re-examine the video to derive the answer, even for queries with directly observable solutions.

\subsection{Answer Leakage Prevention Analysis}
\label{appendix:no_leakage}

To validate the effectiveness of our prompt design, we conducted extensive empirical evaluation with manual verification. As shown in Table~\ref{tab:leakage_validation} (main text), our final prompt design combining structured output constraints with explicit negative prompts achieves zero answer leakage across 200 manually verified teacher-student interactions. The negative prompt component proved particularly effective, reducing leakage from 39.5\% to 5.5\% when used alone, and eliminating it entirely when combined with format constraints. This validation confirms that our prompt design successfully maintains the integrity of the learning process while providing meaningful corrective feedback.

\subsection{Teacher Tool Use Prompt}
\label{appendix:teacher_tools}

The \texttt{TEACHER\_TOOL\_USE\_PROMPT} defines the tools available to the teacher model for detailed video analysis.

\begin{lstlisting}[style=promptstyle, caption={TEACHER\_TOOL\_USE\_PROMPT}]
## Available Tools

You have access to the following tools to examine the video more closely:

### 1. get_frame(frame_index: int)
Retrieves a specific frame from the video for detailed analysis.

### 2. zoom_region(frame_index: int, x1: float, y1: float, x2: float, y2: float)
Zooms into a specific region of a frame. Coordinates are normalized (0-1).

### 3. get_temporal_segment(start_frame: int, end_frame: int, stride: int = 1)
Retrieves a sequence of frames to analyze temporal patterns.

Use these tools when you need to:
- Verify specific visual details mentioned in the student's response
- Identify missed temporal dependencies
- Locate spatial regions containing key evidence
- Validate or refute the student's observations

Call tools in the following format:
TOOL_CALL: tool_name(parameters)

Example:
TOOL_CALL: get_frame(45)
TOOL_CALL: zoom_region(45, 0.3, 0.2, 0.7, 0.6)
\end{lstlisting}

\subsection{Error Classification Summary}
\label{appendix:error_types}

Table~\ref{tab:error_classification} summarizes the error classification system used by the teacher model, along with typical leakage risks and the corresponding mitigation strategies applied in the evidence patch.

\begin{table}[!htb]
\centering
\small
\caption{Error Classification Categories with Leakage Mitigation Strategies}
\label{tab:error_classification}
\resizebox{\columnwidth}{!}{%
\begin{tabular}{p{0.13\linewidth}p{0.28\linewidth}p{0.28\linewidth}p{0.28\linewidth}}
\hline
\textbf{Category} & \textbf{Description} & \textbf{Leakage Risk} & \textbf{Mitigation} \\
\hline
\texttt{temporal} & Wrong temporal order or event sequence & Naming the correct order directly & Redirect to temporal window; use ``before/after'' without specifying the events \\
\texttt{spatial} & Wrong interpretation of frame content & Describing the correct visual content & Point to region coordinates; avoid naming objects/colors/attributes \\
\texttt{misconception} & Misinterpretation of question intent & Restating the question with implicit answer & Clarify the task focus (e.g., ``which action'' vs.\ ``which object'') without revealing the target \\
\hline
\end{tabular}
}
\end{table}
\begin{table*}[!t]
\centering
\caption{Performance comparison of student models using different teacher models. All experiments use Video-R1-SFT-7B as the student model.}
\label{tab:model_comparison}
\resizebox{\textwidth}{!}{%
\renewcommand{\arraystretch}{1.05}
\setlength{\tabcolsep}{4pt}
\begin{tabular}{l cccc cccc}
\toprule
\multirow{2}{*}{\textbf{Model}} & \multicolumn{4}{c}{\textbf{Video Reasoning Benchmark}} & \multicolumn{4}{c}{\textbf{Video General Benchmark}} \\
\cmidrule(lr){2-5} \cmidrule(lr){6-9}
& \textbf{MMVU$\uparrow$} & \textbf{VSI-Bench$\uparrow$} & \textbf{VideoMMMU$\uparrow$} & \textbf{Video-Holmes$\uparrow$} & \textbf{LongVideoBench$\uparrow$} & \textbf{LVBench$\uparrow$} & \textbf{MVBench$\uparrow$} & \textbf{TempCompass$\uparrow$} \\
\midrule
Video-R1-SFT & 61.3 & 31.8 & 47.4 & 34.6 & 47.6 & 30.7 & 59.4 & 69.2 \\
FFR-Qwen3-32B & 67.9 & 38.5 & 50.5 & 47.8 & 52.6 & 34.2 & 68.3 & 72.2 \\
FFR-Qwen3-235B & 68.2 & 38.1 & \textbf{56.5} & 51.6 & 54.2 & 33.9 & 68.8 & 75.2 \\
FFR-GPT-4o & \textbf{69.0} & 38.5 & 55.4 & 49.2 & 53.9 & 36.3 & \textbf{70.0} & 74.9 \\
\rowcolor{cyan!10}
FFR-GLM4.5-V & 68.5 & \textbf{38.9} & 54.6 & \textbf{52.3} & \textbf{55.3} & \textbf{38.1} & 68.8 & \textbf{75.6} \\
\bottomrule
\end{tabular}
}
\end{table*}
\begin{table}[!t]
\centering
\scriptsize
\caption{Training Hyperparameters}
\label{tab:hyperparameters}
\setlength{\tabcolsep}{3pt}
\begin{tabular}{lc|lc}
\hline
\textbf{Hyperparameter} & \textbf{Value} & \textbf{Hyperparameter} & \textbf{Value} \\
\hline
Rollouts ($G$) & 8 & Weight decay & 0.01 \\
Max prompt len. & 16,384 & Batch size (per GPU) & 1 \\
Max completion len. & 1,024 & Grad. accum. steps & 1 \\
Temperature & 1.0 & Max grad. norm & 5.0 \\
Top-p & 0.95 & Training epochs & 1 \\
KL coeff. ($\beta$) & 0.04 & Precision & BF16 \\
Patch tax ($\kappa$) & 0.3 & Attention & FlashAttn-2 \\
Learning rate & 5e-6 & Max image res. & 401,408 px \\
LR scheduler & Cosine & Video frames & 16 \\
\hline
\end{tabular}
\end{table}

\section{Teacher Model Analysis}
\label{appendix:teacher_analysis}

Table~\ref{tab:model_comparison} presents a comprehensive evaluation of how different teacher models influence student performance across video reasoning and general understanding benchmarks. This analysis reveals several key insights about teacher-student dynamics in the FFR framework.

\textbf{Teacher Model Selection Matters.} While all teacher models improve upon the baseline Video-R1-SFT student (61.3 MMVU without teacher), the choice of teacher significantly impacts final performance. GLM-4.5V emerges as the most effective teacher, achieving the best results on 5 out of 8 benchmarks. This superiority likely stems from its strong instruction-following capabilities and robust tool-use integration, enabling more precise error diagnosis and targeted evidence generation.

\textbf{Task-Specific Teacher Advantages.} Different teachers excel at different types of reasoning tasks. GPT-4o shows particular strength in MMVU (69.0), suggesting superior multi-modal understanding, while GLM-4.5V dominates in Video-Holmes (52.3), indicating better temporal reasoning capabilities. This task-specific variation suggests that teacher selection could be optimized based on the target domain.

\textbf{Scaling Effects.} The comparison between Qwen3-32B and Qwen3-235B teachers reveals that simply scaling model size doesn't guarantee better teaching effectiveness. While the 235B model achieves higher VideoMMMU scores (56.5 vs 50.5), the smaller 32B variant performs comparably or better on several other benchmarks. This suggests that teaching ability depends more on model architecture and training methodology than raw parameter count.

\textbf{Teacher Own Benchmark Performance.}
To disentangle the teacher's standalone reasoning ability from its teaching effectiveness, Table~\ref{tab:teacher_own_scores} reports each teacher's own benchmark scores. A stronger standalone teacher does not automatically produce a stronger FFR student. For example, Qwen3-VL-235B achieves the highest average teacher score (68.3), yet the student trained with GLM-4.5V (avg.\ 56.5 in Table~\ref{tab:model_comparison}) is Pareto-optimal overall. This indicates that teaching quality depends not only on the teacher's raw reasoning ability but also on its capacity to generate precise, task-relevant evidence patches without leaking the answer.

\begin{table*}[!htb]
\centering
\caption{Teacher models' own benchmark scores. A stronger standalone teacher does not automatically yield a stronger FFR student (cf.\ Table~\ref{tab:model_comparison}).}
\label{tab:teacher_own_scores}
\resizebox{\textwidth}{!}{%
\renewcommand{\arraystretch}{1.05}
\setlength{\tabcolsep}{4pt}
\begin{tabular}{l cccc cccc c}
\toprule
\multirow{2}{*}{\textbf{Teacher Model}} & \multicolumn{4}{c}{\textbf{Video Reasoning}} & \multicolumn{4}{c}{\textbf{Video General}} & \multirow{2}{*}{\textbf{Avg.}} \\
\cmidrule(lr){2-5} \cmidrule(lr){6-9}
& MMVU & VSI-Bench & VideoMMMU & Video-Holmes & LongVideoBench & LVBench & MVBench & TempCompass & \\
\midrule
GPT-4o & 75.4 & 34.0 & 61.2 & 42.0 & 58.5 & 48.9 & 64.6 & 73.8 & 57.3 \\
GLM-4.5V & 68.7 & 47.5 & 72.4 & 50.3 & 59.8 & 53.8 & 73.0 & 79.2 & 63.1 \\
Qwen3-VL-32B & 75.4 & 61.5 & 71.9 & 42.6 & 61.5 & 63.8 & 76.5 & 79.9 & 66.2 \\
Qwen3-VL-235B & \textbf{75.7} & \textbf{62.7} & \textbf{74.7} & \textbf{45.2} & \textbf{63.2} & \textbf{67.7} & \textbf{76.5} & \textbf{80.6} & \textbf{68.3} \\
\bottomrule
\end{tabular}
}
\end{table*}

\textbf{Repair Success Rate by Error Type.}
To quantify the direct impact of teacher intervention during training, we analyze the \emph{fix success rate}---the fraction of initially incorrect rollouts that become correct after receiving evidence patches. Table~\ref{tab:fix_success_rate} reports the breakdown across error categories on approximately 1{,}000 failed rollouts. Overall, roughly 45\% of failures are repaired after a single teacher intervention, with spatial errors exhibiting the highest recovery rate (${\sim}$52\%), followed by temporal errors (${\sim}$41\%), and misconception errors (${\sim}$38\%). This ordering is consistent with the nature of the evidence: spatial and temporal errors can often be corrected by redirecting attention to specific frames or regions, whereas misconceptions involve deeper misinterpretation of the task intent and are harder to resolve with purely visual guidance. The results confirm that FFR's evidence patches are most effective at recovering missing spatiotemporal cues.

\begin{table}[!htb]
\centering
\small
\caption{Fix success rate by error type on ${\sim}$1{,}000 failed rollouts.}
\label{tab:fix_success_rate}
\begin{tabular}{lccc}
\toprule
\textbf{Error Type} & \textbf{Samples} & \textbf{Fix Rate} & \textbf{Typical Evidence} \\
\midrule
Spatial & ${\sim}$320 & ${\sim}$52\% & Key frames + spatial regions \\
Temporal & ${\sim}$268 & ${\sim}$41\% & Frame ranges + temporal markers \\
Misconception & ${\sim}$412 & ${\sim}$38\% & Error classification + textual guidance \\
\midrule
Overall & ${\sim}$1{,}000 & ${\sim}$45\% & Mixed \\
\bottomrule
\end{tabular}
\end{table}

\section{Experiment Setting Details}
\label{appendix:experiment_setting}

This section provides detailed descriptions of the benchmarks, baselines, and implementation details referenced in Section~\ref{sec:experiment}.

\subsection{Benchmarks}
\label{appendix:benchmarks}

In this work, we evaluate our method on several video reasoning and general video understanding benchmarks to assess its effectiveness and generalization.

\textbf{Video Reasoning Benchmarks.}
We use MMVU~\citep{MMVU}, VSI-Bench~\citep{VSIBench}, VideoMMMU~\citep{VideoMMMU}, and Video-Holmes~\citep{cheng2025video}. These benchmarks cover a variety of tasks, testing models on their ability to answer complex questions by leveraging spatiotemporal context. MMVU focuses on expert-level multi-discipline understanding, VSI-Bench on visual--spatial reasoning, VideoMMMU on model's ability to acquire and apply domain--knowledge, and Video-Holmes on causal and narrative reasoning from multi-segment video clues.

\textbf{General Video Understanding Benchmarks.}
We incorporate LongVideo-Bench~\citep{wu2024longvideobench}, LVBench~\citep{wang2025lvbench}, MVBench~\citep{MVBench}, and TempCompass~\citep{TempCompass}. These benchmarks evaluate models' generalization across diverse video tasks, from long-form video understanding to temporal consistency and motion-based reasoning. LongVideo-Bench tests long-duration video comprehension, LVBench assesses large-scale visual understanding, MVBench challenges multi-modal video understanding and TempCompass examines temporal alignment.

\subsection{Baselines}
\label{appendix:baselines}

We evaluate our method against a range of baselines, categorized as on-policy methods, mixed-policy methods, and tool-use reasoners, as shown in Figure~\textcolor{magenta}{\ref{fig:comp}}.
Specifically, on-policy methods (Figure~\textcolor{magenta}{\ref{fig:comp}} (\textit{a})) include Qwen2.5-VL-7B~\citep{bai2025qwen2}, Video-R1-7B~\citep{VideoR1}, Video-RFT-7B~\citep{VideoRFT}, and Video-ChatR1-7B~\citep{VideoChatR1}.
Mixed-policy methods (Figure~\textcolor{magenta}{\ref{fig:comp}} (\textit{b})) consider AVATAR-7B~\citep{kulkarni2025avatar}.
Tool-use reasoners (Figure~\textcolor{magenta}{\ref{fig:comp}} (\textit{c})) include Pixel-Reasoner~\citep{su2025pixel} and Video-Thinker~\citep{wang2025video}. Specifically, Pixel-Reasoner adopts a two-stage training (SFT-RFT) to learn tool-use abilities, while Video-Thinker is trained for multi-tasking, specifically captioning and temporal grounding.

\subsection{Implementation Details}
\label{appendix:impl_details}

We use Video-R1-7B-SFT and VideoRFT-7B-SFT as our base models and conduct training on 8 NVIDIA A100 GPUs with 80GB memory each. For training efficiency, video inputs are limited to 16 frames, with each frame processed at a resolution of 128$\times$28$\times$28. We train with the FFR framework for 1 epoch, using a learning rate of 5e-6 and generating 8 samples per rollout. The training dataset contains 4,000 samples (corresponding to 500 steps of RL training), sourced from Video-R1, Video-Thinker, and Video-Holmes. Considering computational costs, we select GLM-4.5V as the teacher model. The entire experimental framework is implemented based on R1-V~\citep{chen2025r1v}. During inference, we maintain the same configuration as training, using 16-frame inputs and a maximum pixel resolution of 128$\times$28$\times$28 for model evaluation.

\section{Algorithm Implementation Details}
\label{appendix:implementation}

This section provides implementation details of the FFR framework integrated with GRPO training.

\subsection{Training Pipeline}
\label{appendix:training_pipeline}

Our implementation builds upon the R1-V framework~\citep{chen2025r1v} with modifications to support multi-modal inputs and teacher-guided evidence patches. The training pipeline consists of three main components:

\textbf{1. Rollout Generation.} For each training sample, the student model generates $G$ rollouts (typically $G=8$) using temperature sampling. Each rollout $\tau_i$ consists of the model's reasoning trajectory and final answer. During generation, we maintain visual context (images or videos) alongside text prompts to ensure proper multi-modal reasoning.

\textbf{2. Teacher Intervention.} When a rollout produces an incorrect answer (verified through rule-based matching), the frozen teacher model analyzes the error and provides a minimal evidence patch $c_i$. The teacher uses predefined tools to examine specific frames, temporal segments, or spatial regions, generating targeted guidance without revealing the answer. This intervention is triggered only for incorrect rollouts, creating an adaptive curriculum.

\textbf{3. Second-Round Generation with Evidence.} For incorrect rollouts, the student generates a new response $\tau'_i$ conditioned on both the original input and the teacher's evidence patch. This second-round generation allows the student to correct its reasoning while maintaining on-policy exploration.

\subsection{More Implementation Details}
\label{appendix:implementation_details}

Our implementation integrates the FFR framework into the standard GRPO training loop with minimal modifications. The key implementation choices include:

\textbf{Teacher API Architecture.} The teacher model runs as a separate service to ensure stable performance during distributed training. This decoupling allows the teacher to use different hardware resources and avoids memory conflicts with the student model training.

\textbf{Rollout Selection Strategy.} Among the $G=8$ rollouts per batch, we select the rollout with highest reward as the chosen trajectory $\widehat{\tau}_i$ for policy update. When multiple rollouts achieve the same reward, we prefer those that succeeded without teacher assistance to encourage autonomous reasoning.

\textbf{Synchronization in Distributed Training.} With DeepSpeed ZeRO-3, model parameters are sharded across GPUs. To avoid deadlocks, all ranks must participate in generation even if only some ranks need teacher intervention. We use collective communication primitives to coordinate which rollouts require second-round generation across all ranks.

\subsection{Multi-Modal Processing}
\label{appendix:multimodal}

For video inputs, we sample frames at regular intervals and process them through the vision encoder. Both teacher and student models process 16 frames per video to balance computational cost and temporal coverage. The maximum image resolution is constrained to 401,408 pixels. Visual features are projected and concatenated with text embeddings before the language model backbone.

\subsection{Hyperparameters}
\label{appendix:hyperparameters}
Table~\ref{tab:hyperparameters} summarizes the key hyperparameters used in our experiments.

\section{Additional Experimental Results}
\label{appendix:additional_results}

\subsection{SFT-Teacher vs FFR Comparison}
\label{appendix:sft_comparison}

Table~\ref{tab:sft_comparison_full} presents the full comparison between direct teacher distillation (SFT-Teacher) and FFR across all eight benchmarks (a summary on video reasoning benchmarks is provided in Table~\ref{tab:sft_comparison} in the main text). We use teacher-generated reasoning traces from Qwen3-32B and Qwen3-235B for SFT baselines. The results show that while SFT-Teacher improves over the baseline, FFR significantly outperforms direct distillation, particularly on complex reasoning tasks (+8.4 on VideoMMMU, +5.2 on Video-Holmes compared to SFT-235B). FFR with a 32B teacher already matches SFT with a 235B teacher on overall average (both 54.0), and FFR with GLM-4.5V achieves the best overall performance (56.5). This demonstrates that FFR's targeted intervention mechanism provides more effective learning signals than wholesale knowledge transfer from the teacher.

\begin{table*}[!htb]
\centering
\caption{Full comparison of SFT-Teacher vs.\ FFR under matched teacher models across all benchmarks. FFR consistently outperforms SFT at every scale; FFR with a 32B teacher already matches SFT with a 235B teacher (both Avg: 54.0).}
\label{tab:sft_comparison_full}
\resizebox{\textwidth}{!}{%
\renewcommand{\arraystretch}{1.05}
\setlength{\tabcolsep}{4pt}
\begin{tabular}{l cccc cccc c}
\toprule
\multirow{2}{*}{\textbf{Method}} & \multicolumn{4}{c}{\textbf{Video Reasoning}} & \multicolumn{4}{c}{\textbf{Video General}} & \multirow{2}{*}{\textbf{Avg.}} \\
\cmidrule(lr){2-5} \cmidrule(lr){6-9}
& MMVU & VSI-Bench & VideoMMMU & Video-Holmes & LongVideoBench & LVBench & MVBench & TempCompass & \\
\midrule
Video-R1-SFT (Baseline) & 61.3 & 31.8 & 47.4 & 34.6 & 47.6 & 30.7 & 59.4 & 69.2 & 47.8 \\
\midrule
SFT (Qwen3-32B) & 63.9 & 39.1 & 42.0 & 43.3 & 50.4 & 36.7 & 57.3 & 73.0 & 50.7 \\
SFT (Qwen3-235B) & 67.4 & \textbf{41.9} & 46.2 & 47.1 & 54.5 & \textbf{39.9} & 60.2 & 75.1 & 54.0 \\
\midrule
FFR (Qwen3-32B) & 67.9 & 38.5 & 50.5 & 47.8 & 52.6 & 34.2 & 68.3 & 72.2 & 54.0 \\
FFR (Qwen3-235B) & 68.2 & 38.1 & \textbf{56.5} & 51.6 & 54.2 & 33.9 & \textbf{68.8} & 75.2 & 55.8 \\
\rowcolor{cyan!10}
FFR (GLM-4.5V, Ours) & \textbf{68.5} & 38.9 & 54.6 & \textbf{52.3} & \textbf{55.3} & 38.1 & \textbf{68.8} & \textbf{75.6} & \textbf{56.5} \\
\bottomrule
\end{tabular}
}
\end{table*}

\subsection{Patch Tax $\kappa$ Sensitivity Analysis}
\label{appendix:kappa_sensitivity}

Table~\ref{tab:kappa_sensitivity} presents comprehensive results for different $\kappa$ values across all benchmarks.

\begin{table*}[!htb]
\centering
\caption{Sensitivity analysis of patch tax $\kappa$ across all benchmarks.}
\label{tab:kappa_sensitivity}
\resizebox{\textwidth}{!}{%
\renewcommand{\arraystretch}{1.05}
\begin{tabular}{c cccc cccc c}
\toprule
\multirow{2}{*}{$\kappa$} & \multicolumn{4}{c}{\textbf{Video Reasoning}} & \multicolumn{4}{c}{\textbf{Video General}} & \multirow{2}{*}{\textbf{Avg.}} \\
\cmidrule(lr){2-5} \cmidrule(lr){6-9}
& MMVU & VSI-Bench & VideoMMMU & Video-Holmes & LongVideoBench & LVBench & MVBench & TempCompass & \\
\midrule
0.1 & \textbf{72.7} & 38.7 & 49.0 & 44.3 & 53.6 & \textbf{42.0} & 61.2 & 73.6 & 54.4 \\
0.3 & 68.5 & 38.9 & \textbf{54.6} & \textbf{52.3} & 55.3 & 38.1 & \textbf{68.8} & \textbf{75.6} & \textbf{56.5} \\
0.5 & 69.3 & \textbf{42.8} & 47.3 & 44.2 & 52.4 & 37.6 & 60.6 & 73.8 & 53.5 \\
0.7 & 68.5 & 39.3 & 47.0 & 45.2 & 54.8 & 37.2 & 61.8 & 73.5 & 53.4 \\
1.0 & 70.6 & 40.4 & 47.9 & 43.0 & \textbf{57.8} & 35.8 & 60.8 & 75.4 & 54.0 \\
\bottomrule
\end{tabular}
}
\end{table*}

The results demonstrate that FFR is robust to different $\kappa$ values, with all configurations outperforming the baseline. $\kappa=0.3$ achieves the best overall performance (56.5\%), particularly on complex reasoning benchmarks. Too small a penalty ($\kappa{=}0.1$) leads to teacher over-reliance, while larger values ($\kappa \geq 0.5$) excessively discount teacher contributions.

\subsection{Detailed Case Study}
\label{appendix:case_study}

Table~\ref{tab:case_study_detail} provides a detailed breakdown of the case study shown in Figure~\ref{fig:case_study}, illustrating how FFR's evidence patches enable reasoning correction without answer leakage. This case is drawn from a real training sample in the STAR dataset.

\begin{table*}[!htb]
\centering
\caption{Detailed case study of FFR intervention on a temporal reasoning task from STAR dataset.}
\label{tab:case_study_detail}
\resizebox{\textwidth}{!}{%
\begin{tabular}{p{3cm}p{13cm}}
\toprule
\textbf{Component} & \textbf{Content} \\
\midrule
\textbf{Source} & STAR Dataset, Video: 5QW1X.mp4 \\
\midrule
\textbf{Question} & Which object was put down by the person? \\
\midrule
\textbf{Options} & A. The clothes \quad B. The book \quad C. The shoe \quad D. The dish \\
\midrule
\textbf{Ground Truth} & B (The book) \\
\midrule
\textbf{Student's Original Response} & \textcolor{red}{\ding{55}} Answer: A (The clothes) \\
& \textit{Reasoning:} ``I see clothes in the scene...'' \\
& $\rightarrow$ Focused on visible clothes, missed the actual put-down action \\
\midrule
\textbf{Error Classification} & \texttt{temporal\_error}, \texttt{spatial\_error} \\
\midrule
\textbf{Evidence Patch} & \\
\quad Key Frames & [13, 14, 15] \\
\quad Temporal Markers & ``when the person's hand releases the object'', ``moment of placement on surface'' \\
\quad Guidance & ``Track hand movements and which object is placed down.'' \\
\midrule
\textbf{Student's Corrected Response} & \textcolor{green!60!black}{\ding{51}} Answer: B (The book) \\
& \textit{Reasoning:} ``At frames 8--10, the person's hand releases the book onto the surface. Clothes remain in place throughout.'' \\
\midrule
\textbf{Why This Patch Works} & \\
\quad \textcolor{green!60!black}{\ding{51}} Minimal \& Sufficient & Points to specific frames [13, 14, 15]; provides temporal cues (hand release); guides attention without revealing answer \\
\quad \textcolor{red}{\ding{55}} Does NOT & State ``the book was put down''; reveal the correct answer is B; describe exact frame contents \\
\bottomrule
\end{tabular}
}
\end{table*}

This case exemplifies the core design principle of FFR: the evidence patch redirects the student's attention to the relevant visual evidence without short-circuiting the reasoning process. The student must still observe the specified frames, track the hand movements, and reason about which object transitions from being held to being placed down.

\subsection{Generalization Across Student Scales and Architectures}
\label{appendix:generalization}

A natural question is whether FFR generalizes beyond the 7B Qwen2.5-VL students used in the main experiments. We evaluate two complementary axes: \emph{model scale} (a smaller 3B student) and \emph{model architecture} (the Qwen3-VL family).

\textbf{Smaller Student: Qwen2.5-VL-3B.}
Table~\ref{tab:3b_results} reports results on Qwen2.5-VL-3B, with and without an SFT warm-up, following the same training protocol as the 7B experiments. FFR improves the base 3B model by +10.0 in average accuracy even without SFT. Under the SFT-then-FFR setting that mirrors our 7B protocol, FFR further lifts the average from 43.0 to 45.7, with notable gains on VideoMMMU (+7.1) and LongVideoBench (+6.8). This confirms that the benefit of observation-level teacher intervention is not limited to a particular model scale.

\begin{table*}[!htb]
\centering
\caption{FFR on Qwen2.5-VL-3B across all benchmarks. FFR provides substantial gains even at the 3B scale.}
\label{tab:3b_results}
\resizebox{\textwidth}{!}{%
\renewcommand{\arraystretch}{1.05}
\setlength{\tabcolsep}{4pt}
\begin{tabular}{l cccc cccc c}
\toprule
\multirow{2}{*}{\textbf{Model}} & \multicolumn{4}{c}{\textbf{Video Reasoning}} & \multicolumn{4}{c}{\textbf{Video General}} & \multirow{2}{*}{\textbf{Avg.}} \\
\cmidrule(lr){2-5} \cmidrule(lr){6-9}
& MMVU & VSI-Bench & VideoMMMU & Video-Holmes & LongVideoBench & LVBench & MVBench & TempCompass & \\
\midrule
Qwen2.5-VL-3B (base) & 44.8 & 24.8 & 29.8 & 28.7 & 39.5 & 29.0 & 48.1 & 31.8 & 34.6 \\
\quad $+$ FFR & 53.3 & \textbf{37.7} & 35.2 & 35.0 & 42.3 & 31.1 & \textbf{57.5} & \textbf{64.8} & 44.6 \\
\midrule
Qwen2.5-VL-3B-SFT & 53.8 & 35.5 & 33.3 & 32.4 & 41.8 & 29.3 & 53.8 & 64.3 & 43.0 \\
\rowcolor{cyan!10}
\quad $+$ FFR & \textbf{57.9} & 31.5 & \textbf{40.4} & \textbf{35.6} & \textbf{48.6} & \textbf{31.8} & 55.9 & 64.1 & \textbf{45.7} \\
\bottomrule
\end{tabular}
}
\end{table*}

\textbf{Different Architecture: Qwen3-VL-8B.}
To test whether FFR transfers beyond the Qwen2.5-VL family, we apply FFR to Qwen3-VL-8B using Qwen3-VL-32B as the teacher. Table~\ref{tab:qwen3_results} shows that despite the much stronger baseline of Qwen3-VL-8B (avg.\ 62.5), FFR still improves the average to 65.1, clearly outperforming vanilla GRPO (62.8). The gains are most pronounced on VideoMMMU ($+$4.2) and MMVU ($+$3.0). On tasks requiring complex spatiotemporal reasoning, the FFR-trained 8B student even surpasses its 32B teacher: Video-Holmes (48.7 vs.\ 42.6) and LongVideoBench (68.4 vs.\ 61.5). This asymmetry is expected because the teacher's role in FFR is \emph{diagnostic}---localizing errors and providing evidence patches---which does not require strong end-task accuracy. The student then converts these diagnostic signals into improved answer-generation strategies through on-policy exploration, a process that can yield higher task accuracy than the teacher's own direct answers.

\begin{table*}[!htb]
\centering
\caption{FFR on Qwen3-VL-8B (teacher: Qwen3-VL-32B). The FFR-trained 8B student surpasses its 32B teacher on video reasoning benchmarks.}
\label{tab:qwen3_results}
\resizebox{\textwidth}{!}{%
\renewcommand{\arraystretch}{1.05}
\setlength{\tabcolsep}{4pt}
\begin{tabular}{l cccc cccc c}
\toprule
\multirow{2}{*}{\textbf{Model}} & \multicolumn{4}{c}{\textbf{Video Reasoning}} & \multicolumn{4}{c}{\textbf{Video General}} & \multirow{2}{*}{\textbf{Avg.}} \\
\cmidrule(lr){2-5} \cmidrule(lr){6-9}
& MMVU & VSI-Bench & VideoMMMU & Video-Holmes & LongVideoBench & LVBench & MVBench & TempCompass & \\
\midrule
Qwen3-VL-8B (base) & 70.2 & 58.9 & 57.7 & 45.3 & 66.6 & 55.7 & 69.5 & 76.4 & 62.5 \\
\quad $+$ GRPO & 71.2 & 59.5 & 57.3 & 46.4 & 66.4 & 55.2 & 70.3 & 76.2 & 62.8 \\
\rowcolor{cyan!10}
\quad $+$ FFR & \textbf{73.2} & \textbf{61.2} & \textbf{61.9} & \textbf{48.7} & \textbf{68.4} & \textbf{57.1} & \textbf{72.0} & \textbf{78.2} & \textbf{65.1} \\
\midrule
\rowcolor{gray!10}
\textit{Qwen3-VL-32B (teacher)} & \textit{75.4} & \textit{61.5} & \textit{71.9} & \textit{42.6} & \textit{61.5} & \textit{63.8} & \textit{76.5} & \textit{79.9} & \textit{66.2} \\
\rowcolor{gray!10}
\textit{Qwen3-VL-235B} & \textit{75.7} & \textit{62.7} & \textit{74.7} & \textit{45.2} & \textit{63.2} & \textit{67.7} & \textit{76.5} & \textit{80.6} & \textit{68.3} \\
\bottomrule
\end{tabular}
}
\end{table*}

Together with the 7B results in the main text, these experiments confirm that FFR's benefits hold across three model scales (3B, 7B, 8B) and two distinct architecture families (Qwen2.5-VL and Qwen3-VL).

\subsection{Inference-Time Frame Scaling}
\label{appendix:frame_scaling}

Our training protocol uses 16 frames per video. To evaluate whether the learned reasoning patterns transfer to richer visual inputs at inference time, we test the FFR checkpoint (Video-R1-SFT + FFR, trained at 16 frames) at 32 and 64 frames \emph{without any additional training}. Table~\ref{tab:frame_scaling} reports the results.

\begin{table*}[!htb]
\centering
\caption{Inference-time frame scaling. The FFR checkpoint trained at 16 frames generalizes to 32 and 64 frames with consistent improvements, particularly on long-video benchmarks.}
\label{tab:frame_scaling}
\resizebox{\textwidth}{!}{%
\renewcommand{\arraystretch}{1.05}
\setlength{\tabcolsep}{4pt}
\begin{tabular}{c cccc cccc c}
\toprule
\multirow{2}{*}{\textbf{Frames}} & \multicolumn{4}{c}{\textbf{Video Reasoning}} & \multicolumn{4}{c}{\textbf{Video General}} & \multirow{2}{*}{\textbf{Avg.}} \\
\cmidrule(lr){2-5} \cmidrule(lr){6-9}
& MMVU & VSI-Bench & VideoMMMU & Video-Holmes & LongVideoBench & LVBench & MVBench & TempCompass & \\
\midrule
16 & 68.5 & 38.9 & 54.6 & 52.3 & 55.3 & 38.1 & \textbf{68.8} & 75.6 & 56.5 \\
32 & 69.8 & 39.4 & 55.4 & \textbf{52.9} & 57.2 & 41.7 & 61.3 & 76.9 & 56.8 \\
64 & \textbf{69.4} & \textbf{41.8} & \textbf{56.1} & 51.1 & \textbf{58.4} & \textbf{42.0} & 64.6 & \textbf{78.3} & \textbf{57.7} \\
\bottomrule
\end{tabular}
}
\end{table*}

The model generalizes well to increased frame budgets: the overall average improves from 56.5 (16 frames) to 57.7 (64 frames). Long-video benchmarks benefit most (LongVideoBench $+$3.1, LVBench $+$3.9), as the additional temporal coverage better matches these tasks' demands. Short-video benchmarks such as MVBench show some fluctuation due to redundant frames, but 6 out of 8 benchmarks improve at 64 frames. These results confirm that the spatiotemporal reasoning patterns learned by FFR at 16 frames transfer effectively to richer visual inputs without retraining.

\subsection{Robustness Under Inference Seeds}
\label{appendix:seed_robustness}

To assess the stability of FFR's improvements, we evaluate the same trained checkpoint (Video-R1-SFT + FFR) across three different inference seeds. Table~\ref{tab:seed_robustness} reports per-benchmark results and summary statistics.

\begin{table*}[!htb]
\centering
\caption{Robustness analysis across inference seeds. The trained FFR checkpoint exhibits low variance ($\pm$0.4 overall), confirming that the observed gains are not artifacts of sampling noise.}
\label{tab:seed_robustness}
\resizebox{\textwidth}{!}{%
\renewcommand{\arraystretch}{1.05}
\setlength{\tabcolsep}{4pt}
\begin{tabular}{l cccc cccc c}
\toprule
\multirow{2}{*}{\textbf{Seed}} & \multicolumn{4}{c}{\textbf{Video Reasoning}} & \multicolumn{4}{c}{\textbf{Video General}} & \multirow{2}{*}{\textbf{Avg.}} \\
\cmidrule(lr){2-5} \cmidrule(lr){6-9}
& MMVU & VSI-Bench & VideoMMMU & Video-Holmes & LongVideoBench & LVBench & MVBench & TempCompass & \\
\midrule
Seed 42 & 68.5 & 38.9 & 54.6 & 52.3 & 55.3 & 38.1 & 68.8 & 75.6 & 56.5 \\
Seed 123 & 69.0 & 38.9 & 54.5 & 52.1 & 56.0 & 39.6 & 68.6 & 75.5 & 56.8 \\
Seed 456 & 68.9 & 38.2 & 55.4 & 51.2 & 55.4 & 36.4 & 68.7 & 75.6 & 56.2 \\
\midrule
\rowcolor{cyan!10}
Mean $\pm$ Std & 68.8$\pm$0.2 & 38.6$\pm$0.3 & 54.8$\pm$0.4 & 51.9$\pm$0.5 & 55.6$\pm$0.3 & 38.0$\pm$1.3 & 68.7$\pm$0.1 & 75.6$\pm$0.1 & 56.5$\pm$0.4 \\
\bottomrule
\end{tabular}
}
\end{table*}

The overall standard deviation across seeds is only $\pm$0.4, with per-benchmark variation below 1.0 on 7 out of 8 benchmarks (LVBench exhibits slightly higher variance at $\pm$1.3, likely due to its smaller test set). These results confirm that FFR's gains are robust and not sensitive to inference-time sampling noise.

\end{document}